\documentclass[12pt, article]{imsart}

\RequirePackage[OT1]{fontenc}
\RequirePackage{amsthm,amsmath} 
\usepackage{natbib}
\RequirePackage[colorlinks,citecolor=blue,urlcolor=blue]{hyperref}
\RequirePackage{hypernat}
\usepackage{amssymb}
\usepackage{multirow}
\usepackage{capt-of}
\usepackage{amsmath}
\usepackage{algorithm}
\usepackage[noend]{algpseudocode}
\usepackage{graphicx}
\usepackage{graphicx}
\usepackage{booktabs,caption,fixltx2e}
\usepackage[flushleft]{threeparttable}
\usepackage{subcaption}
\usepackage{relsize}

\usepackage[markup=underlined]{changes}

\usepackage[compact]{titlesec}
\titlespacing{\section}{1pt}{*0.5}{*0.1}
\titlespacing{\subsection}{0.7pt}{*0.3}{*0.1}
\titlespacing{\subsubsection}{0pt}{*0}{*0}

\usepackage{xcolor}

\newcommand{\modified }{\color{black}}     
\makeatletter
\def\BState{\State\hskip-\ALG@thistlm}
\makeatother

\renewcommand{\baselinestretch}{1.3}
\makeatletter
\@addtoreset{equation}{section}
\renewcommand{\theequation}{\thesection.\arabic{equation}}
\makeatother

\relax

\newcommand{\bay}{\begin{array}}
\newcommand{\eay}{\end{array}}

\newcommand{\bqa}{\begin{eqnarray*}}
\newcommand{\eqa}{\end{eqnarray*}}
\newcommand{\bqan}{\begin{eqnarray}}
\newcommand{\eqan}{\end{eqnarray}}
\newcommand{\bqt}{\begin{quote}}
\newcommand{\eqt}{\end{quote}}
\newcommand{\bt}{\begin{tabbing}}
\newcommand{\et}{\end{tabbing}}
\newcommand{\bit}{\begin{itemize}}
\newcommand{\eit}{\end{itemize}}
\newcommand{\ben}{\begin{enumerate}}
\newcommand{\een}{\end{enumerate}}
\newcommand{\beq}{\begin{equation}}
\newcommand{\eeq}{\end{equation}}
\newcommand{\bdefi}{\begin{definition}}
\newcommand{\edefi}{\end{definition}}
\newcommand{\bpro}{\begin{proposition}}
\newcommand{\epro}{\end{proposition}}
\newcommand{\blem}{\begin{lemma}}
\newcommand{\elem}{\end{lemma}}
\newcommand{\bth}{\begin{theorem}}
\newcommand{\bco}{\begin{corollary}}
\newcommand{\eco}{\end{corollary}}
\newcommand{\bdes}{\begin{description}}
\newcommand{\edes}{\end{description}}

\newtheorem{definition}{Definition}[section]
\newtheorem{proposition}[definition]{Proposition}
\newtheorem{lemma}[definition]{Lemma}
\newtheorem{theorem}[definition]{Theorem}
\newtheorem{corollary}[definition]{Corollary}

\renewcommand{\baselinestretch}{1.2} 

\newcommand{\cv}{cross-validation }
\newcommand{\thp}{thresholding parameter }

\begin{document}
	
\thispagestyle{empty}		
	\begin{frontmatter}
		\title{
		Different thresholding methods on 
 \protect{Nearest Shrunken Centroid} algorithm 
}
		
		\vglue 5mm

		\begin{aug}
			\author{\fnms{Mohammad Omar} \snm{Sahtout$^1$}\ead[label=e1]{mosahtout@ucdavis.edu}}
and
\author{\fnms{Haiyan} \snm{Wang$^{2a}$}\corref{} \ead[label=e2]{hwang@ksu.edu}}
and
\author{\fnms{Santosh} \snm{Ghimire$^3$} \ead[label=e3]{santoshghimire@ioe.edu.np} }
			\affiliation{{\small Department of Statistics, Kansas State
					University, 101 Dickens Hall, Manhattan, KS 66506}
			}

			\address{
               $^1$  Department of Statistics, University of California, Davis\\
               $^2$ Department of Statistics, Kansas State University\\
               $^3$ Department of Applied Sciences and Chemical Engineering, Pulchowk Campus, Tribhuvan University
			}

			\runauthor{Running Head   
}
		\end{aug}
		\footnotetext{\textit{$^*$ Running head: Different thresholding on PAM}}
		
		\footnotetext{\textit{$^{a}$ Author to whom correspondence may be
				addressed. Email: hwang@ksu.edu}}
		
		\runtitle{Effect of different thresholding on PAM}

		\medskip
\end{frontmatter}
\newpage		

\setcounter{page}{1}
\begin{frontmatter}

\title{ 
		Different thresholding methods on
 nearest shrunken centroid algorithm }	

\vspace{0.5in}
		\begin{abstract}
This article considers the impact of different thresholding methods to the Nearest Shrunken Centroid algorithm, which is popularly referred as the Prediction Analysis of Microarrays (PAM) for high-dimensional classification. PAM uses soft thresholding to achieve high computational efficiency and high classification accuracy but in the price
 of retaining too many features. When applied to microarray human cancers, PAM selected 2611 features on average from 10 multi-class datasets. Such a large number of features make
 it difficult to perform follow up study.
 One reason behind this problem is the soft thresholding, which is known to produce biased parameter estimate in regression analysis.
  In this article,
we extend the PAM algorithm with two other thresholding methods, hard and order thresholding, and a
deep search algorithm to achieve better thresholding parameter estimate.
The modified algorithms are extensively tested and compared to the original one based
 on real data and Monte Carlo
 studies.
 In general, the modification
not only gave better cancer status prediction accuracy, but also resulted in
more parsimonious models with significantly smaller number of features.
\end{abstract}

	\end{frontmatter}

\section{Introduction}
Cancer classification is one of many fields where high-dimensional data are abundantly available. Molecular data such as gene expression can be used to help classify different cancer types and
subtypes. A key challenge for such application
is the
large number of predictors compared to the relatively small sample size. Typically, model based
 variable selection and classification require to estimate a large number of unknown
parameters which is difficult when the sample size is small. Recent efforts have been focused mostly on variable screening to filter
out irrelevant variables or using penalization methods to shrink small parameter estimates
toward zero. Thresholding is one of the techniques used for this purpose.



\label{makereference2.1}

Three thresholding methods are popular in classification models due to their computational efficiency to handle ultra high dimensional features and low sample sizes.
Soft Thresholding was developed to denoise the noisy signal via shrinking toward zero \citep{Bickel1983}. Hard Thresholding was meant to give better reproduction of signal height but with discontinuities \citep{Donoho.Johnstone1994}.
Order Thresholding estimates signals if the observed value is above certain percentile \citep{Kim.Akritas2010}.
Various studies have established that thresholding methods perform better than their non-thresholded counterparts in
differentiating sparse signals from noise provided that
  the thresholding parameter is chosen appropriately.
\cite{Donoho.Johnstone1994}
 suggested
 to use a universal thresholding parameter $(2 \log n )^{1/2}$ for both soft and hard thresholding.
 \cite{Fan1996} suggested taking the thresholding parameter to be $(2 \log(n a_n) )^{1/2}$, with $a_n=c(\log n)^{-d}$ for some positive constants $c$ and $d$.
On the other hand, cross-validation is used in \cite{Tibshirani.et.al2002} as a data-driven rule to select the soft thresholding parameter for their nearest shrunken centroids classifier. \cite{Kim.Akritas2010} recommended to estimate the optimal parameter for order thresholding with $(\log n)^{3/2}$ in their simulations.

Each of the aforementioned thresholding and parameter estimation was developed for their specific domain of application, which is not necessarily a classifier.
 Here we would like to assess computationally how these different thresholding and parameter estimations would work with one specific classifier.   Some examples of binary classifiers are support vector machine (SVM)  \citep{Chang.Lin2011} and its variants such as SCAD-SVM \citep{Zhang.et.al2006}, Features Annealed Independence Rule  \citep{Fan.Fan2008}, Binary Matrix Shuffling Filter \citep{Zhang.Wang2012} and Top Scoring Pairs family \citep{Tan.et.al2005}.
  Though there are many classifiers available for binary data, not many are suited for multi-class data.
 Popularly used multi-class classifiers include the Naive Bayes classifiers, Nearest Shrunken Centroid \citep{Tibshirani.et.al2002}, k-nearest neighbor, and lasso or elastic net logistic regression \citep{Friedman10regularizationpaths}. Ensemble of different or same classifiers on different subspaces could also lead to classifiers that can handle multi-class data. Well known examples are Random Forest \citep{Breiman2001}, 
 Bagging \citep{Breiman1996}, 
 Boosting \citep{Freund1997}, 
 and extreme gradient boosting \citep{Chen.Guestrin2016}.
 These ensemble classifiers generally are much more computationally extensive.

 In this article, we focus on the Nearest Shrunken Centroid 
 classifier {\modified PAM}
 because it works for both binary and multi-class problems and is computationally efficient. PAM often can give better performance than regression methods in high dimensional classification when sample sizes are small to moderate. A supporting study is \cite{Ng.Jordan2002} who concluded that regression based discriminative model requires a linear number of training samples in number of parameters to converge to its asymptotic error while a generative classifier may require only a logarithmic number of training samples to converge to its asymptotic error. PAM is a generative classifier.
 Thresholding method
and the parameter estimate are two major components of the PAM which govern the accuracy of the classifier.
In this article we consider hard and order thresholdings
 to improve the performance of the  classifier. Moreover, a deep search algorithm for better
thresholding parameter estimate will be introduced. We examine three different thresholding
parameter estimates and compare them with the thresholding parameters obtained from cross-validation. 

The rest of the article is organized as follows. In section \ref{makereference_sec2}, we {\modified start by presenting the original PAM classifier, then we} present the
 modified PAM algorithm by considering hard and order thresholdings. Numerical comparisons of the three thresholding methods are given in section \ref{makereference_sec3}. {\modified Finally, we end with discussions in Section \ref{makereference2.2} for comparison with earlier findings and an overall summary of this work. 
 }




\section{Materials and methods
}
\label{makereference_sec2}

\subsection{Notation for the original PAM classifier}
\label{makereference3}
Given a set of $n$ training samples from $K$ different classes and each is a vector with $p$ feature variables, the single entry $x_{ij}$ represents the observed value for the $i^{th}$ feature variable from the $j^{th}$ sample, and $y_j$ represents the class label for sample $j$.
 As the original PAM was described in the analysis of microarray setting, the features were referred as genes and observed values are expression values of genes. In this article, {\bf we interchangeably use feature variables and genes to refer to the covariates}.

 Assume that the classes are labeled 1 through $K$, such that $y_j\in \{1, 2,\ldots, K\}$. Let $n_k$ denote the number of samples from class $k$ and $C_k$ be the set of indices for those samples.
Then, the centroid of the $k^{th}$ class can be written as a function of the t statistic $d_{ik}$ as:
\bqa \label{PAM3} \overline{x}_i^{(k)} = \overline{x}_i+m_k(s_i+s_0)d_{ik}.\eqa
where  $\overline{x}_i=\sum_{j=1}^n{x_{ij}/n},$
$ d_{ik}=\frac{\overline{x}_i^{(k)}-\overline{x}_i}{m_k(s_i+s_0)}, $
 $ s_i^2=\frac{\sum_k{\sum_{j \in C_k}{(x_{ij}-\overline{x}_i^{(k)})^2}}}{n-K}, $
 $m_k=\sqrt{1/n_k+1/n},$ and $s_0$ is a small quantity to guard against zero for the denominator.

The PAM classifier shrinks
the $d_{ik}$ values to $d_{ik}^{ \backprime}$ with the soft thresholding. They are then used to define the new shrunken centroid as
\bqan \label{PAM4} \overline{x}_i^{\backprime(k)} = \overline{x}_i+m_k(s_i+s_0)d_{ik}^{ \backprime}. \eqan
The resulting shrunken centroids in (\ref{PAM4}) is used for classifying any new sample, say $x^*=(x_1^*, x_2^*, \ldots , x_p^*)$, by first computing the discriminant score for each class using
\bqan \label{PAM6} \delta_k(x^*)=\sum_{i=1}^p{\frac{(x_i^*-\overline{x}_i^{\backprime(k)})^2}{(s_i+s_0)^2}}-2\log{\pi_k}, \eqan
where $\pi_k$ is the class prior probability.
 Then, $x^*$ will be classified to the class having the smallest discriminant score.

 When examining the details of how PAM chose the thresholding parameter for Leukemia2 dataset \citep{Armstrong.et.al2002}, 
 we observed that the number of genes survived soft thresholding corresponding  to the smallest cross-validation error could differ by thousands from that corresponding to the second smallest cross-validation error. On the other hand, the smallest and 2nd smallest cross-validation errors might differ by one misclassified sample. This could be a potential problem of the thresholding parameter estimate in PAM. Illustration of this potential problem using Leukemia2 dataset is presented in Table \ref{Table}. Moreover PAM selects too many features which makes it difficult to perform follow up experiments in cancer studies. We believe one of the reasons for this drawback is the soft thresholding method PAM used. In a regression set up, such as lasso or regularized logistic regression \citep{Friedman10regularizationpaths},
the soft thresholding on the ordinary (or iteratively reweighted) least square estimate with certain thresholding parameter achieves the $L_1$ penalty on the regression parameters. Even though soft thresholding provides computational advantage, the solution is biased.

Put Table \ref{Table} about here.

Considering these issues, 
we experiment two ways of improving the PAM algorithm. One way is to replace the soft thresholding used in the PAM algorithm by either hard or order thresholding. The second way is to give a better estimate of the thresholding parameter. We will
provide an algorithm that performs a deep search for selecting the optimal thresholding parameter.

\subsection{Nearest shrunken centroids classification with different thresholding methods}
\label{makereference3.1.1}
Let $d_{ik}^{ \backprime}$ denote the thresholded value of the test statistic $d_{ik}$
using soft thresholding \bqan \label{PAM53} d_{ik}'=sgn(d_{ik})(|d_{ik}|-\Delta _S)_+. \eqan
%
%
To alter the PAM algorithm, we replace the soft thresholding in (\ref{PAM53}) by the hard thresholding
\bqan \label{HT3} d_{ik}^{\backprime\backprime}=d_{ik} I\{|d_{ik}|>\Delta_H\}, \eqan and order thresholding
\bqan \label{OT3} d_{ik}^{\backprime\backprime\backprime}=\left\{\begin{array}{ll} d_{ik} & \mbox{if rank}(|d_{ik}|)>n-\Delta _O \\  0   & \mbox{otherwise} \\
  \end{array} \right.\eqan
 As described in Section \ref{makereference3}, we obtain the shrunken centroids $\overline{x}_i^{\backprime\backprime(k)}$, $\overline{x}_i^{\backprime\backprime\backprime(k)} $ and discriminant scores  $\delta_k^{\backprime\backprime}$, $ \delta_k^{\backprime\backprime\backprime}$ for hard and order thresholdings respectively. Then a new sample $z$ will be classified to the class with the smallest discriminant score.

\textbf{Estimation of the thresholding parameter} \hspace{.5cm}
Assume we start with a set of $m$ thresholding  parameter
values $\Theta_0=\{\theta_{01}, \ldots, \theta_{0m}\}$.
 Without loss of
generality, we assume that $\theta_{01}, \ldots, \theta_{0m}$ are
arranged in an increasing order.
We repeatedly shrink the search range to
find the value of the best thresholding parameter. Specifically, let
$Err(\theta_{0i})$ represent the cross-validation error of the
algorithm when $\theta_{0i}$ is the thresholding parameter, where
$i=1, \ldots, m$. Define $\tau_1=\underset{1 \leq i \leq
m}{\mbox{argmin}}\{Err(\theta_{0i}), i=1,\ldots, m\}$ to be the index of
the thresholding parameter value whose corresponding
cross-validation error $Err(\theta_{0\tau_{1}})$ is the smallest
among all parameter values in set $\Theta_0$. That is, inequality
$Err(\theta_{0k}) \ge Err(\theta_{0\tau_{1}})$ holds for all $k\ne
\tau_1$. Then the optimal thresholding parameter is
in the interval $(\theta_{0,\tau_{1}-1}, \theta_{0,\tau_{1}+1})$. We
then consider a second set of \thp values $\Theta_1=\{\theta_{11},
\ldots, \theta_{1m}\}$ evenly spaced in the interval
$(\theta_{0,\tau_{1}-1}, \theta_{0,\tau_{1}+1})$. The parameter
value in $\Theta_1$ that has the smallest \cv error is then
identified. Denote it as $\theta_{1,\tau_{2}}$. This leads to a even
smaller interval $(\theta_{1,\tau_{2}-1}, \theta_{1,\tau_{2}+1})$
for further search. The process is repeated and a sequence of
intervals $(\theta_{i-1,\tau_{i}-1}, \theta_{i-1,\tau_i+1})$ is
obtained for $i=1, 2, \ldots$ . The search will be terminated when
the number of variables surviving the thresholding remains unchanged
for all parameters in an interval. Beyond repeatedly narrowing the search grid of the thresholding parameter, the other thing to consider is the thresholding value corresponding to the second smallest cross-validation error. Below is an algorithm to refine the thresholding parameter estimate.

\textbf{Deep Search algorithm }
\bit
\item[1.] Start by searching within the $m$ thresholding values (m=30 default) to find the thresholding values corresponding to the smallest and 2nd smallest cross-validation (CV) error (i.e. $\theta_{0\tau}$ and $\theta_{0\nu}$).
    \begin{itemize}
    \item In case of more than one thresholding values with the same CV error, choose the one with the smallest number of selected variables.
    \item Set the temporary further search location as $\theta_{temp} =\theta_{0\tau}$.
    \end{itemize}
 \item [2.]The thresholding value corresponding to the 2nd smallest CV error ($\theta_{0\nu}$) can be assigned to $\theta_{temp}$ in our algorithm if both conditions in 2a and 2b are satisfied.
    \begin{itemize}
    \item[2a.] The difference between the smallest and the 2nd smallest CV error does not differ by more than one misclassified sample (i.e. $Err(\theta_{0\nu})-Err(\theta_{0\tau})\leq1$).
    \item[2b.] The number of variables survived thresholding corresponding to the second smallest CV error ($g_{0\nu}$) is either
        \begin{itemize}
        \item less than half of that for the thresholding value with the smallest CV error (i.e. $2 g_{0\nu} < g_{0\tau}$),\\ or
        \item 2,000 less than that for the thresholding value with the smallest CV error (i.e. $g_{0\tau} - g_{0\nu} > 2000$).
        \end{itemize}
    \end{itemize}
After locating this initial thresholding value ($\theta_{temp}$), the next process will be to deeply search the neighborhood of $\theta_{temp}$ for another possible thresholding value with smaller CV error. Record the index $\ell$ in $\Theta_0$ such that $\theta_{temp}=\theta_{0\ell}$.
 \item[3.] To identify the neighboring interval that will be investigated, consider both sides of the selected thresholding value ($\theta_{temp}$). That is, both intervals ($\theta_{0,\ell-1}$, $\theta_{0\ell}$) and ($\theta_{0\ell}$, $\theta_{0,\ell+1}$).
    \begin{itemize}
    \item In case the selected thresholding value in step 2 is a boundary value (i.e. $\ell = 1$ or $\ell= m$).
        \begin{itemize}
        \item If $\ell=1$, just consider the right side of the selected thresholding value (i.e. interval ($\theta_{0\ell}$, $\theta_{0\ell+1}$)).
        \end{itemize}
    \item The following two conditions specify which interval to perform the deep search:
        \begin{itemize}
        \item Only perform the deep search on the interval ($\theta_{0\ell}$, $\theta_{0,\ell+1}$) if the difference in number of selected variables is more than one variable (i.e. $g_{0\ell} - g_{0,\ell+1} > 1$).
        \item Only perform the deep search on the interval ($\theta_{0\ell-1}$, $\theta_{0,\ell}$) if the difference in number of selected variables is less than $m$ (i.e. $g_{0,\ell-1} - g_{0\ell} < m$).
        \item If both  $g_{0\ell} - g_{0,\ell+1} > 1$ and $g_{0,\ell-1} - g_{0\ell} < m$ conditions are satisfied, perform the deep search in ($\theta_{0\ell-1}$, $\theta_{0,\ell+1}$).\\
        \end{itemize}
    \end{itemize}
After deciding on which interval to refine the search ($\theta_{0,\ell-1}$, $\theta_{0,\ell+1}$), ($\theta_{0,\ell -1}$, $\theta_{0\ell}$), or ($\theta_{0\ell}$, $\theta_{0,\ell+1}$):
\item[4.]  Now consider a second set of \thp values $\Theta_1=\{\theta_{11}, \ldots, \theta_{1k}\}$ evenly spaced in the selected interval from the previous step.
    \begin{itemize}
    \item The number of thresholding values $k$ is the minimum between $m$ and the difference between the number of variables that correspond to the lower and upper bounds of the interval. For example, if the selected interval is ($\theta_{0\ell}$, $\theta_{0,\ell+1}$), then the number of thresholding values to be considered is $k=\min(m, g_{0\ell} - g_{0,\ell+1})$.
    \end{itemize}
\item[5.] Run cross-validation to obtain the CV errors for the set of selected thresholding values from the previous step.
\item[6.] If $k>0$, repeat steps 1 to 5 with the parameter values in $\Theta_1$. Otherwise, report the optimal thresholding value as the most recently obtained $\theta_{temp}$.
\eit
The above algorithm basically starts with a set of thresholding values and use cross-validation to obtain the initial best thresholding value which has the smallest cross-validation error. Then around the neighborhood, find an even better one, which has error close to the best but not in the expense of too many variables.

\subsection{Materials}
\label{makereference3.2}
Ten multi-class gene expression data sets for human cancers were investigated in this study and are listed in Table \ref{tab1}. These data sets were kindly provided by the authors of \cite{Tan.et.al2005}.
The number of classes in those data sets ranges from 3 to 14 and the number of genes ranges from
2308 to 16063.

Put Table \ref{tab1} about here.


  For easier discussion, from now on we will refer to the hard thresholded PAM algorithm by HTh, the order thresholded PAM algorithm by OTh, and the soft thresholded PAM algorithm (the original PAM) by STh.

 The R software 
 was used for programming of those three PAM algorithms. For
 STh, we mainly used functions from the pamr package that was developed by the authors of \cite{Tibshirani.et.al2002}. To perform deep search for STh,
  we start with 30 initial thresholding values 
 and refine the neighborhood of the value that has the smallest cross-validation error following the search procedure described at the end of Section \ref{makereference3.1.1}. Specifically, we first identify a shorter interval and evenly re-split this interval into 30 values. Then we calculate their cross-validation error for each value. This process will continue until we reach the thresholding value with the smallest cross-validation error. After determining the shrinkage parameter using cross-validation, the pamr.train function is used to build the classifier with the informative genes that survived the thresholding. Then the model is used to classify the class label of each test sample by applying the method of nearest centroid classification using the pamr.predict function.

 For the HTh and OTh algorithms, we wrote our own functions to calculate the class centroids, to perform cross-validation, and to predict the class label for the test samples. The refining process is also implemented in our code for these two algorithms. In all three algorithms the number of folds for the cross-validation is set to be 10 unless some class sample size is less than 10. In the later case, the fold is set to be the smallest class size.

 STh, HTh, and OTh use the smallest cross-validation error for the thresholding parameter estimate. The deep search algorithm in Section \ref{makereference3.1.1} results in possibly different parameter estimate. We refer to these algorithms using soft, hard, and order thresholding along with deep search algorithm for parameter estimate as STh2, HTh2, and OTh2, respectively.\\\\
\textbf{Comparison metric} \hspace{.5cm}
In binary classification problems, multiple metrics can be used for comparison. In the case of at least 3 classes, proportion of correctly classified samples or (misclassified samples) is typically used in the literature as the comparison metric. When discussion is within the same dataset, the number of misclassified test samples by different methods can also be used. We will use the test error in our comparison. It is defined as the percent of misclassification error, which is equal to the number of misclassified test samples divided by the total number of test samples. We will also compare the number of variables selected in each method and refer them as informative genes.

\section{Results}
\label{makereference_sec3}
\subsection{Performance of STh, OTh, and HTh}
\label{makereference3.2.1}

In this section, we discuss the performance of the three algorithms using the 10 multi-class human cancers data sets. In all that follows, our reported misclassification error refers to the percentage of misclassified test samples.  Since random partition of the training data in cross-validation could lead to different estimated thresholding parameter and hence possibly a different test error, we repeated this process 100 times for each dataset.
\subsubsection{Comparison of performance on each dataset}
\label{makereference3.2.1.1}

We start by discussing the results of each dataset individually.
For better visualization of our comparison, in {\modified Figure \ref{I1} (and Figures \ref{I2} to \ref{I6} in the Supplementary Material)} we plotted the test errors of the STh against the test errors of both OTh and HTh.
We computed {\bf the number of times out of 100 runs that the OTh has less test error than the STh and this proportion is given in the plots as $P(Err_o<Err_s)$}. Similarly, $P(Err_h<Err_s)$ and $P(Err_o<Err_h)$ are given in the plots with their meaning accordingly defined. Below those plots we reported the mean, median, and standard error of the different algorithms based on the 100 runs. The average number of informative genes for each algorithm is also reported.

We begin with the small round blue cell tumors (SRBCT) dataset
analysis. Figure \ref{I1}(a) shows the scatter plot
of the STh test error versus HTh
and OTh test errors from 100 runs. We can see that only one sample
out of the 20 test samples was misclassified for all three
methods in all 100 runs, except for one run for the OTh that has misclassified 5 samples.
So for this dataset the three methods are almost
equivalent with a 5\% test error. However, the average number of
informative genes used by the OTh is equal to one third of the
number used in the STh. The total number of genes in this dataset is 2308 genes. The average number of
informative genes used by the OTh is 1.4\% of the total number of genes. HTh used 4 more genes on average than
the OTh.

Put Figure \ref{I1} about here.

Figure \ref{I1} (b) displays the result for the Breast cancer
dataset analysis. The probabilities given in the plot
represents the proportion of times out of 100 runs that the test
error for one thresholding method is smaller than another. The OTh has the smallest test error and has the
smallest average number of informative genes. The HTh has the
highest mean test error, but similar median test error to the STh.
The STh selected the highest number of genes again. It used more
than 3300 (36\% of the total number of genes), while the OTh only
used 7.4\% of total number of genes to achieve even better performance. The STh has
the smallest standard error for the mean test error. The OTh has higher
 standard error and the HTh has the largest standard error.

Figure \ref{I2}(a) {\modified in the Supplementary Material} presents the result from the analysis of the Cancers
dataset. This dataset contains different types of cancer
samples: prostate, breast, lung, ovary, colorectum, kidney, liver,
pancreas, bladder/ureter, and gastroesophagus. For this dataset the
STh has the best performance in that it has the smallest mean test
error, standard error for mean test error, and average number of
informative genes. In fact, almost all test errors of STh in 100
runs reached the smallest of the three methods except in one run, in
which the HTh has one less misclassified sample than STh. All three
methods used more than 1000 genes, but the OTh used 81 genes less than the HTh.
The OTh and the HTh have the same median test
error. The percentages of identified informative genes by the three
methods are 8.9\% with the STh, 11.8\% with the OTh, and 12.4\% with
the HTh method.



Moving to Figure \ref{I2}(b) {\modified in the Supplementary Material}, this analysis is for the diffuse
large B-cell lymphoma (DLBCL) dataset. In this case the OTh and the
HTh always have less test errors than that for the STh in all 100
runs. Both OTh and HTh have zero median test error. Even though the
HTh had a better mean test error and standard error, the OTh had the
smallest average number of selected genes. There is a very big
difference in the number of selected genes between the OTh and the
STh methods, as OTh selected 360 genes while STh selected 3483
genes.

On the other hand, we see the opposite result with the GCM dataset analysis given
 in Figure \ref{I3} {\modified in the Supplementary Material}. The OTh and the
HTh always have larger test errors than that for the STh in all 100
runs. This dataset is a collection of
samples from 14 common human tumor types and it has the largest
number of genes. In this analysis all three algorithms had
the worst test error rate among all data sets in this study. In addition,
all three methods used more than 2000 genes. The median test error
was 43.48\% for the STh and 52.17\% for both OTh and HTh. The OTh
has the smallest standard error (0.1) while the HTh has the largest (0.27) standard error. The average number of selected genes ranged from 2010 for the STh to 3716 genes for the HTh method.



Figures $\ref{I4}-\ref{I6}$ {\modified in the Supplementary Material} are for Leukemia
cancer data sets. Even though all of them are for the same cancer,
the results based on the three data sets are very different. This
might be due to the following reasons: (1) The number of classes in
these three datasets are different. There are 3 classes in Leukemia1
and Leukemia2 but there are 7 classes in Leukemia3 data. (2) The
training sample sizes are different (38, 57, 215 for Leukemia1,
Leukemia2, and Leukemia3 respectively.) (3) The genes and the number
of genes in the three data sets are different. Leukemia1 data used a
much earlier version of Affymetrix GeneChip array that has 7129
genes. Leukemia2 and Leukemia3 used later versions of Affymetrix
GeneChip array(s), one with 12582 genes and the other one with 12558
genes. In terms of accuracy, STh appears to be the best method out of
the three for two of the data sets but has the worst performance in
the remaining dataset. In terms of the average number of informative
genes, however, the STh has the worst performance in two out of the
three data sets. It is interesting to see that the number of genes
that survived thresholding with the STh method show a clear
association with the version of Affymetrix GeneChip array. In the
earlier version (i.e. Leukemia1 data) STh has 111 genes survived
while in the later version(s) more than 5300 genes survived
thresholding.

%
%

For the Leukemia1 dataset, Figure \ref{I4}(a) {\modified in the Supplementary Material} summarizes the
result of this analysis. Here the STh has 3\% mean test error and
111 average number of selected genes. Both values are less than
those for either OTh or HTh. In all 100 runs, STh has the smallest
test error among all 3 methods. The HTh and OTh have comparable
performance in test errors but the OTh used less number of
informative genes.

Figure \ref{I4}(b) {\modified in the Supplementary Material} for Leukemia2 dataset shows STh has the
worst performance among the three methods in that it not only has
the largest average and median test errors but also has trouble in finding informative genes. The final model of STh kept on average 5389 genes,
which is 16 times more than that used by OTh. The OTh has the
smallest average number of selected genes (327). There is also a big
difference in the number of selected genes for the HTh (1492) and
STh (5389). The HTh has similar median test error of 6.67\% to that
for the OTh but smaller standard error. The median test error for
the STh is 20\%.


The analysis for the third Leukemia cancer dataset (Figure
\ref{I5} {\modified in the Supplementary Material}) shows that the STh has the least mean test
error (3.01\%) but with a very large average number of selected
genes, 4606.
OTh has an average of 5.07\% test error with an average
number of genes being 1020. HTh has mean test error of 4.73\% with
average number of informative genes being 2070. On average the HTh
has 
comparable
mean test error to the OTh but with the price of using
 1050 more genes on average.

The last two data sets are for Lung cancer. The analysis of Lung1 dataset in Figure \ref{I6}(a) {\modified in the Supplementary Material} shows that OTh and HTh have equivalent performance in terms of the test error but OTh used less genes. The STh has the smallest average number of selected genes in this case. Figure \ref{I6}(b) {\modified in the Supplementary Material} presents the analysis for the Lung2 dataset with best performance achieved by OTh followed by the STh. In this case the OTh
classified all test samples correctly in all 100 runs. The STh had
the smallest average number of informative genes, 1911. The OTh used 2106 genes and the HTh used the highest number of genes (3610).\\


\subsubsection{Overall comparison based on all ten data sets}
\label{makereference3.2.1.2}
It can be seen from the previous section that none of the three
algorithms (STh, OTh, and HTh) is absolutely the best across all ten
data sets. In this section, we combine the results from
different data sets and provide an overall comparison. Specifically,
we have the average percentage of misclassification errors for each
method based on 100 runs for each cancer dataset. We also have the
average number of informative genes from the 100 runs per method and
dataset combination. Since the percent of misclassification errors
are mostly small while the numbers of informative genes from
different methods have drastically different ranges, a nonparametric
approach without the assumption of constant variance and normality
is more meaningful than a parametric method.

We will use the Sum of Ranking Difference (SRD) proposed in
\cite{Heberger2010} and \cite{Heberger2011} to do the comparison.
This method assumes that there is a golden standard. In our setting, the golden
standard can be set to be the best performance out of all methods
being compared.

Put Table \ref{tab8} about here.

\begin{figure}[H]
\caption{Overall comparison of STh, HTh, and OTh bases on SRD. {\modified The x-axis and the left y-axis represents SRD values scaled to between 0 and 100; the right y-axis gives the relative frequencies for the theoretical distribution. The XX1, Med, and XX19 mark the 5\%, 50\%, and 95\% percentiles. }}
 \label{I7}
 \centering
\footnotesize
\begin{tabular}{cc}
\includegraphics[height=5cm,width=.45\linewidth, angle=0]{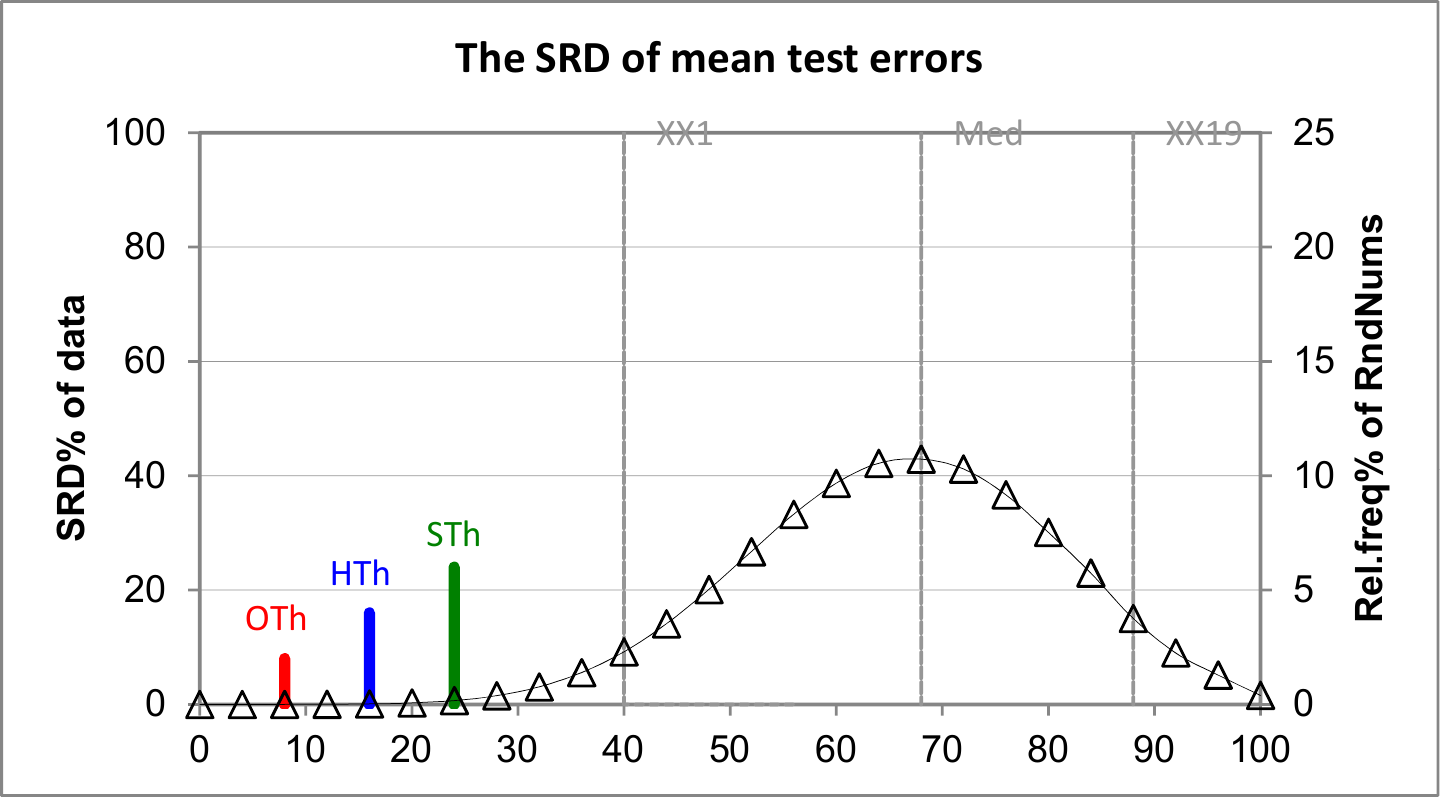}
&
\includegraphics[height=5cm,width=.45\linewidth, angle=0]{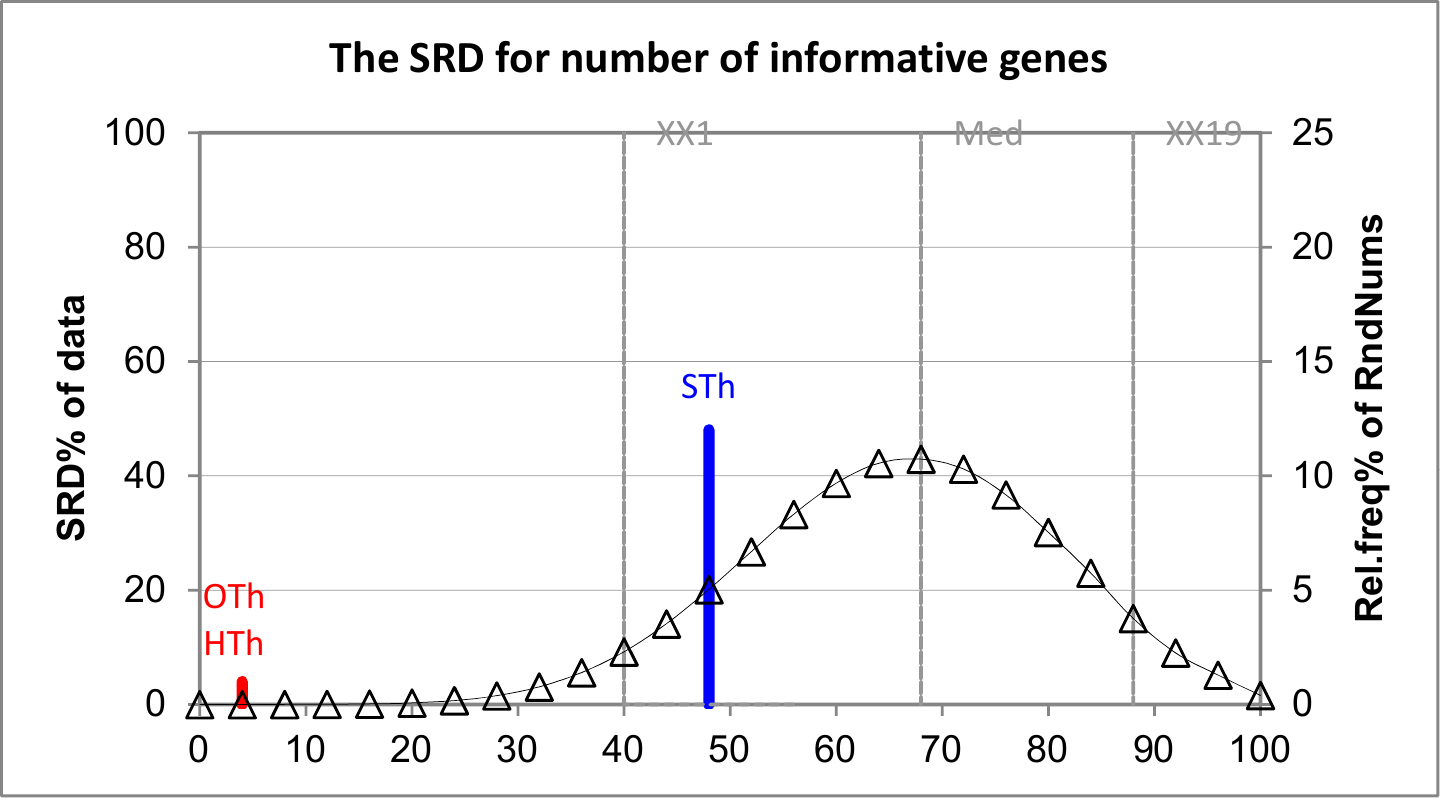}
 \\
(a) The SRD comparison of mean test
errors  & (b) The SRD comparison for the number of\\
  & informative genes \\
\end{tabular}
\end{figure}

We first applied this method to the mean test errors to compare the three algorithms. We assume the golden standard  for each dataset to be the minimum of the mean test error across three algorithms. {\modified The CRRN-DNA software was  downloaded from the link given in  \cite{Heberger2011} to calculate SRD and its theoretical distribution.} Table \ref{tab8} shows the SRD calculations based on the mean test errors from 100 runs on each method for all three methods. The minimums of the mean test error across three algorithms for each dataset, as our golden standard, are shown in the third column. {\modified The ranks of the mean test errors of each algorithm on different data sets are
given in the columns labeled as ``rank".} The ranks for the OTh algorithm are similar to the ideal ranking except for three data sets. {\modified The absolute difference between each algorithm's
ranks and the ideal rank are in the ``diff" columns.} The sum of those differences for each algorithm is the sum rank difference and is given in the last row of the table. This result is presented in Figure \ref{I7} (a). The OTh has the smallest sum rank difference (4); the HTh is in the middle with sum rank difference (8); and the STh has the largest sum rank difference value (12). This means that the OTh is the closest to the golden standard, the minimum mean test error. Therefore, according to the SRD method the OTh is the best algorithm in terms of the test error. The HTh is second and the STh has the largest mean test error.

{\modified Beyond showing the relative position of the SRD values of the three algorithms, Figure \ref{I7} also shows the theoretical distribution of the SRDs under the null hypothesis that the given SRD for an algorithm can be derived from random ranking. The theoretical distributions
of the SRD values are generated for random numbers (for sample size less than 14) or approximated using normal distribution for large number of samples (more than 13)
(see \cite{Heberger2011}). In our case we are evaluating the different algorithms using 10 cancer samples.  With sample size of 10, the theoretical distribution is based on permutation distribution. The distribution's $5^{th}$ and $95^{th}$ percentiles are labeled as XX1 and XX19 in Figure \ref{I7}. All three algorithms' SRD values in Figure \ref{I7} panel (a) fall below the XX1 vertical line. That is, the probability that these three solutions were derived randomly is much less than 5\%. }


Put Figure \ref{I7} about here.
%

Put Table \ref{tab3} about here.

Beside the prediction accuracy of classifiers, identifying informative genes is very important for the researcher. This importance comes from the need to reduce the large number of irrelevant genes such that biologically important genes can be identified for further experimentation in followup studies. Hence, the number of informative genes identified by each thresholding method is another comparison criterion often used in the literature. In Table \ref{tab3} we listed the average number of informative genes selected over 100 runs of each algorithms for each
dataset. OTh has the smallest overall average number of informative genes across all ten data sets (see the bottom row of Table \ref{tab3}). In addition, the OTh was the most consistent, in terms of the number of informative genes, compared to the other two algorithms. Its standard error of the average number of informative genes ranged from 0.9 to 47.5. While for the HTh it ranged from 1 to 258 and for the STh it ranged from 6 to 414.9. Even though the STh method identified a reasonable number of informative genes in some cases, it resulted in very large numbers in four datasets: Breast, DLBCL, Leukemia2, and Leukemia3. The HTh never had the minimum average number of informative genes. It had either the middle value or the largest value. In terms of overall average across all data sets as shown in the bottom row of Table \ref{tab3}, HTh is in between OTh and STh. We applied also the SRD approach to the average number of informative genes from 100 runs for each dataset to compare those three thresholding methods. As shown in Figure \ref{I7}(b) the OTh and the HTh have tied SRD value that is much smaller than {\modified the 5\% significance level. The SRD value for the STh is not significantly different from random ranking.}

\subsection{Performance of STh2, OTh2, and HTh2}
\label{makereference3.2.2}
In this section, we discuss the performance of STh2, OTh2, and HTh2, the improved versions of the STh, HTh, and OTh  respectively.
Again we use deep search algorithm (\ref{makereference3.1.1}) for selecting optimal thresholding parameter. Our analysis for this section is also for the 10 multi-class human cancers datasets listed in Table \ref{tab1}. In all that follows, our reported misclassification error refers to the percentage of misclassified test samples.

\begin{figure}[H]
\caption{Overall comparison of STh2, HTh2, and OTh2 bases on SRD. {\modified The x-axis and the left y-axis represents SRD values scaled to between 0 and 100; the right y-axis gives the relative frequencies for the theoretical distribution. The XX1, Med, and XX19 mark the 5\%, 50\%, and 95\% percentiles. }}
 \label{I8}
 \centering
\footnotesize
\begin{tabular}{cc}
\includegraphics[height=5cm,width=.45\linewidth, angle=0]{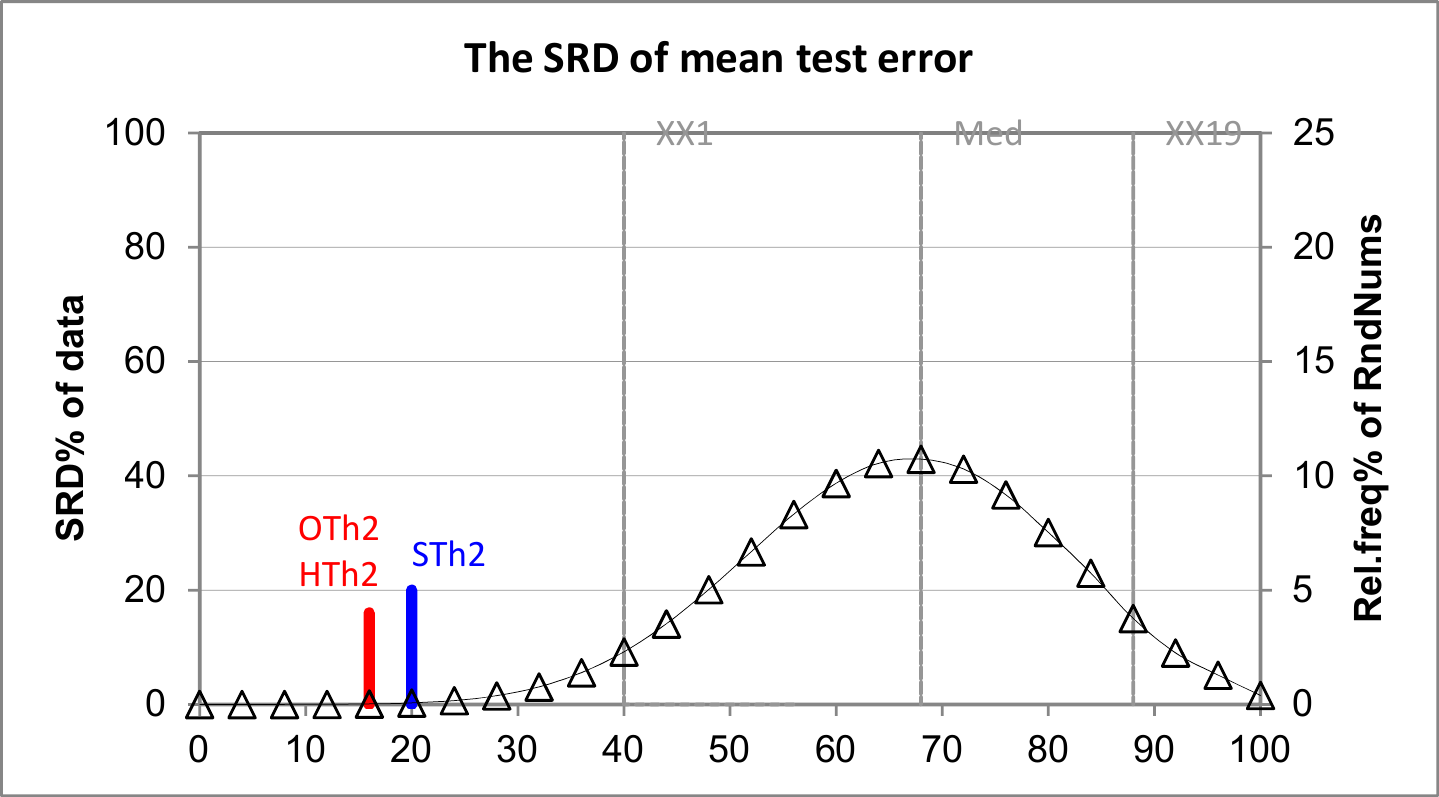}
&
\includegraphics[height=5cm,width=.45\linewidth, angle=0]{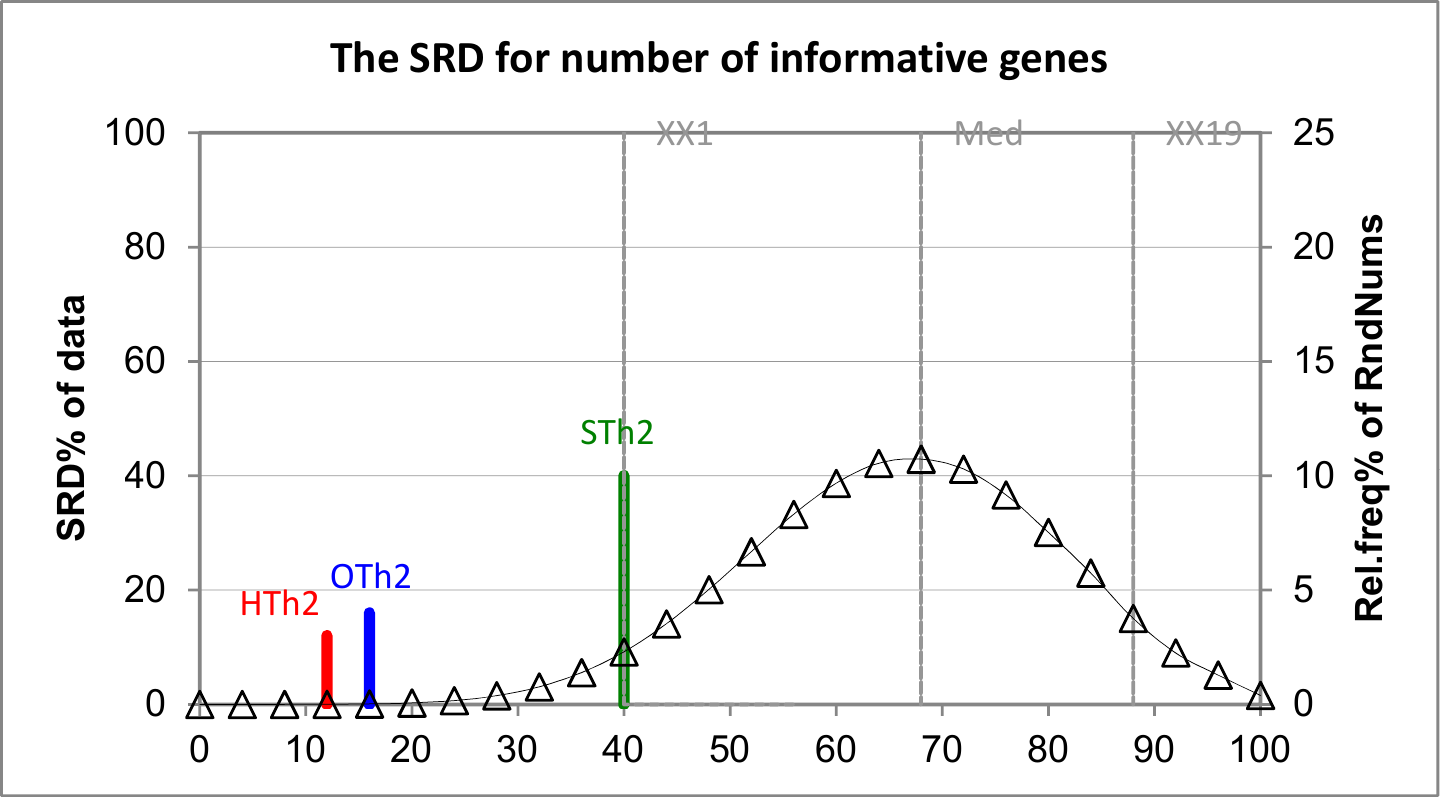}
 \\
(a) The SRD comparison of mean test
errors  & (b) The SRD comparison for the number of \\
  &  informative genes \\
\end{tabular}
\end{figure}

The Sum of Ranking Difference (SRD) will be used again in this section to compare the different algorithms. We will start by comparing the performance of the three algorithms that use the deep search: the STh2, HTh2, and OTh2. Figure \ref{I8}(a) presents the results for the SRD of mean test errors for these three algorithms. The golden standard for the SRD method is assumed to be the minimums of the mean test error across three algorithms for each dataset. The OTh2 and HTh2 have the same sum rank difference and it is smaller than that for the {\modified STh2}. This means that they are closer to the minimum mean test error than the {\modified STh2}. {\modified All three algorithms' SRD values are less than the 5\% significance level indicated by the XX1 position in the plot. Hence the rankings of these three algorithms are all significantly different from random ranking.} Therefore, according to the SRD method the OTh2 and HTh2 are better algorithms in terms of the test error than the STh2. Comparing these results to those of STh, HTh, and OTh in Figure \ref{I7}(a), it is clear that using the deep search algorithm reduced the SRD value for the STh2. Moreover, the differences between the SRD for the three algorithms that use the deep search are smaller.

Put Figure \ref{I8} about here.

The SRD results for the number of informative genes are presented in Figure \ref{I8}(b). The golden standard for the SRD method is assumed to be the minimums for the number of informative genes across three algorithms for each dataset. The HTh2 has the smallest sum rank difference, which means that it is the closest to the minimum number of informative genes. Therefore, according to the SRD method the HTh2 is the best algorithm in terms of the number of informative genes. The OTh2 is in the middle and STh2 still has much larger SRD value. {\modified Both HTh2 and OTh2 have  significantly lower SRD values than random ranking while STh2 overlaps with the 5\% significance level.} Comparing these results to those of STh, HTh, and OTh in Figure \ref{I7}(b), we also noticed that using the deep search algorithm reduced the SRD value for the STh2 from that of STh.

\begin{figure}[H]
\caption{Overall comparison of STh, HTh, OTh, STh2, HTh2, and OTh2 bases on SRD. {\modified The x-axis and the left y-axis represents SRD values scaled to between 0 and 100; the right y-axis gives the relative frequencies for the theoretical distribution. The XX1, Med, and XX19 mark the 5\%, 50\%, and 95\% percentiles. }}
 \label{I9}
 \centering
\footnotesize
\begin{tabular}{cc}
\includegraphics[height=5cm,width=.45\linewidth, angle=0]{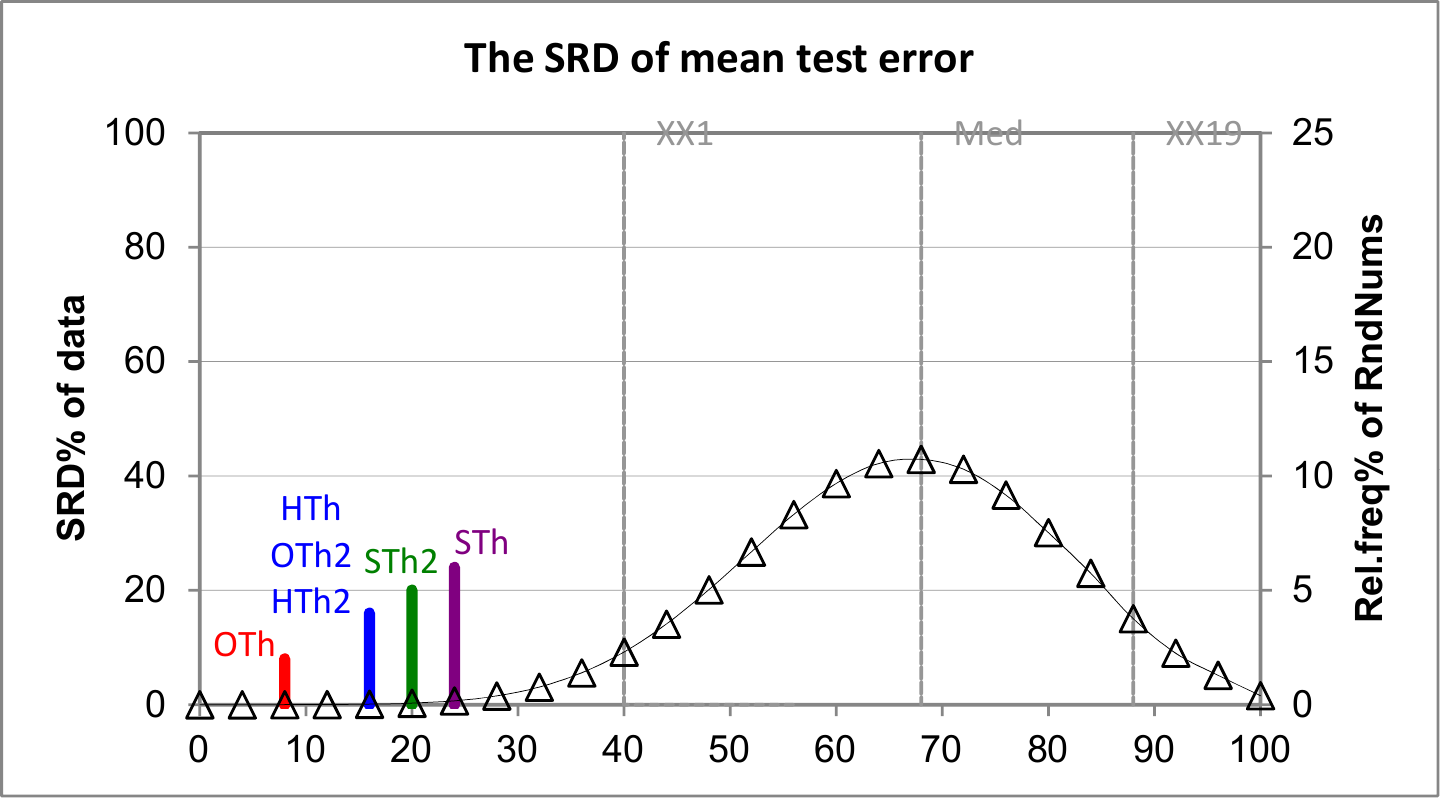}
&
\includegraphics[height=5cm,width=.45\linewidth, angle=0]{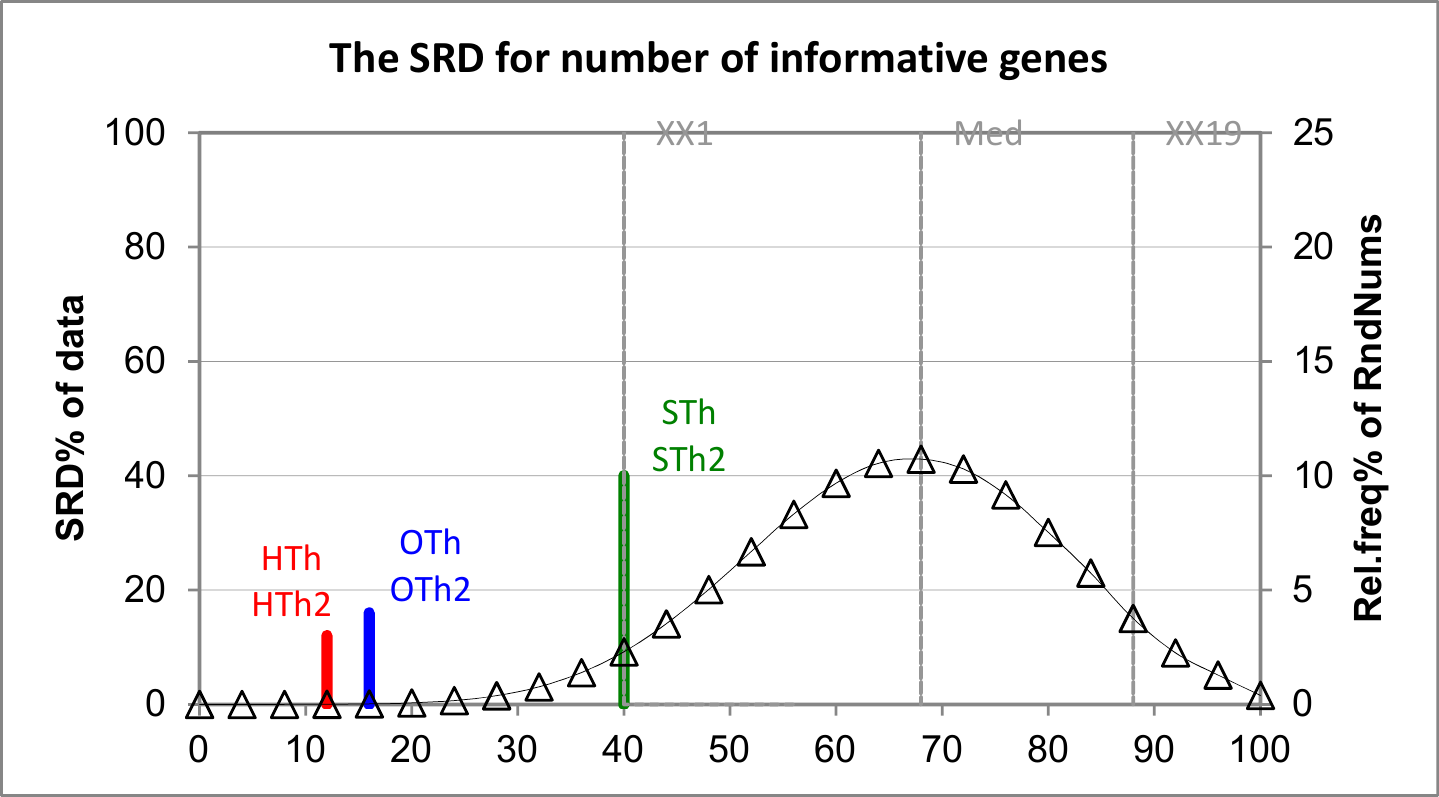}
 \\
(a) The SRD comparison of mean test
errors  & (b) The SRD comparison for the number of \\
  &  informative genes \\
\end{tabular}
\end{figure}

Next we compare the performance of all six algorithms STh, OTh, HTh, STh2, OTh2, and HTh2. Figure \ref{I9}(a) presents the results for the SRD of mean test errors for these six algorithms. Among all six algorithms, the OTh has the smallest sum rank difference, which means that it is the best algorithm in terms of the test error. The OTh2, HTh2 and HTh have tied second SRD value. The STh2 and STh have larger SRD values. Therefore, according to the SRD method the OTh is the best algorithm and the STh is the worst algorithm in terms of the test error if all six algorithms were compared. {\modified The ranking results are significant in that the SRD values of all algorithms are below 5\% significance level. }

Put Figure \ref{I9} about here.

The results of the SRD method for the number of informative genes for all six algorithms are presented in Figure \ref{I9}(b). The interesting observation in this figure is that each one of the deep search algorithms has the same SRD value as its counterpart. The HTh2 and its counterpart HTh have the smallest sum rank difference. The OTh2 with OTh are in the middle and STh2 with STh have much larger SRD value. {\modified The SRD values of HTh, HTh2, OTh, and OTh2 are all significantly below the 5\% significance level. So their rankings are significantly different from random ranking. The STh and STh2's rankings are a marginal case as their SRD value overlaps with the 5\% vertical line. }

For a closer view of the results of the algorithms that use the deep search and to compare them with their counterpart algorithms ({\modified{STh}}, OTh, and HTh), Table \ref{tab4} presents the mean test errors and the average number of informative genes based on 100 runs for each of the six algorithms. The average number of selected genes by the STh2 was reduced from those by STh for all datasets. The mean test errors for the STh2 stayed almost the same as those for the STh except for the Leukemia2 dataset, for which STh2 has 7\% less mean test error than its counterpart STh. For the Leukemia2 dataset, the average number of selected genes for the STh2 is 2236, while it was 5389 for those of STh. The average number of selected genes for the HTh2 was reduced from those of HTh for all data sets except for both Cancers and Leukemia1 data sets. The difference in mean test errors between HTh2 and its counterpart HTh is not more than 2\% except for the Leukemia2 dataset (about 4\%). In addition, the difference in mean test errors between OTh2 and its counterpart OTh is not more than 2\% except for the Leukemia2 dataset (about 5\%). Even though the difference in average number of selected genes for the OTh2 and OTh are not as large as those between STh2 and STh or HTh2 and HTh, there is still obvious reduction except for GCM, Leukemia1, and Lung1 data sets.

In conclusion, the deep search algorithm results in significant decrease in the number of selected genes for each method, while it kept the mean test errors barely changed. That is, the algorithms with deep search and their counterpart without deep search have similar test errors in that the difference in the test errors is no more than 2\%.

Put Table \ref{tab4} about here.

\section{Discussion}

\label{makereference2.2}

In this work, we introduced different approaches to modify
Nearest Shrunken Centroid 
 algorithm in order to alleviate the problem of retaining
 too many features in the original PAM algorithm.
 Our first approach was to replace the soft thresholding by
hard or order thresholding so that the thresholding parameter estimate is less biased.
 Beyond thresholding methods, choosing the parameter value out of a fixed set of values in the original
PAM algorithm could have a potential problem of missing the optimal thresholding parameter. This happens because only a finite number of thresholding values were evaluated in the original
PAM algorithm while the parameter space is continuous. The risk
will increase when considering the smallest cross-validation error as
a single selection {\modified criterion}.
To overcome this problem
we take into consideration of how likely the smallest cross-validation error approximates the true error by comparing the 2nd smallest cross-validation error to the smallest cross-validation error. In addition, we implemented a deep search algorithm to repeatedly refine the neighborhood of the initially selected
thresholding parameter value to reach a better parameter estimate.

 \subsection{{\modified  Comparison with earlier findings in scope of application and choice of threshold} }
{\modified PAM has soft thresholding integrated in its algorithm. Hard thresholding or order thresholding was never used in PAM in the literature or in real practice. In fact, the order thresholding in \cite{Kim.Akritas2010} was established under the setting of a sequence of independent Gaussian observations $X_i \sim N(\theta_i, 1)$ and to test $H_0: \theta_i =0$. It was then applied to high dimensional one-way ANOVA from observations from normal distribution.
We modified PAM to allow hard thresholding or order thresholding to be used. The resulting algorithm is actually different from the original PAM even though we still refer them as PAM.

The soft thresholding was introduced by \cite{Bickel1983} for multivariate normal decision theory. The hard thresholding is simply the ``keep or discard" rule frequently used in model selection in regression analysis. Under the setting of decision theory where the observed data is equal to the signal plus Gaussian white noise, \cite{Donoho.Johnstone1994} theoretically proved that using soft or hard thresholding on the coordinate estimates exhibits the same asymptotic performance. They established that the universal order of the upper bound for least squares error loss is $2\log(n)$ times the sum of the ideal risk and the mean squared loss for estimating one parameter unbiasedly when assuming the oracle is known, where $n$ is the sample size. The results were then applied to nonlinear function estimation using adaptive wavelet shrinkage and piecewise polynomial. They concluded that wavelet  selection  using an oracle can closely mimic piecewise polynomial fitting using an oracle and piecewise polynomials fit are not more powerful than wavelets fit in nonlinear regression. \cite{Donoho.Johnstone1994} also concluded that variable-knot  spline fits when equipped with an oracle to select the knots, are not dramatically more powerful than selective wavelet reconstruction with an oracle. Theoretical properties with known oracle is nice. However, in real practice, the oracle is not accessible. The aforementioned asymptotic properties may not be achieved when denied access to an oracle and forced to rely on data alone.

After \cite{Donoho.Johnstone1994} laid the theoretical ground, soft thresholding and hard thresholding have been used increasingly more and more in applications. For example, \cite{Fan1996} used both soft and hard thresholding to test whether a sequence of independent Gaussian variables with variance one has mean 0 and also applied to goodness of fit test.  \cite{Johnstone.Silverman2004} gave a number of additional applications of thresholding including image processing, model selection, and data mining. The most use of these thresholding rules occur in regression or likelihood estimation problems because when a thresholding rule is applied to the model parameters, it is equivalent to an $L_p$ penalty on the parameters. The hard thresholding corresponds to $L_0$ penalty on the parameter values while the soft thresholding corresponds to $L_1$ penalty. Since the $L_1$ penalty preserves sparsity and convexity which makes parameter estimation easy with a closed form formula, the soft thresholding is much more widely used than the hard thresholding even though $L_1$ penalty leads to biased solutions (see \cite{Fan.Li2006} for the form of the bias). This also explains why the original PAM was implemented with soft thresholding.

We conducted comparison of the data driven thresholding parameter estimation via cross-validation to general inference-based thresholding parameter estimates that were recommended in literature. {\modified Even though there is theoretical support for the universal threshold $(2 \log n)^{1/2}$ for both soft and hard thresholding by \cite{Donoho.Johnstone1994}, and the threshold  $(2 \log(n a_n))^{1/2}$ by \cite{Fan1996} for hard thresholding, with $a_n = c(\log n)-d$ for some positive constants $c$ and $d$, and
$(\log n)^{3/2}$ by \cite{Kim.Akritas2010} for order thresholding, these optimal thresholds require large $n$.}
The overall comparison, using our real data analysis, was
in favor of the thresholding parameter estimates obtained from cross-validation. This is likely due to the fact that the sample sizes are mostly not large.

}

\subsection{{\modified Ranking of different algorithms } }

{\modified Beyond modifying the original PAM with different thresholding methods, we made comprehensive comparison of the resulting algorithms using ten data sets.
Comparing performance using one or two data sets can easily give a conclusion. Using ten data sets, however, requires an effective tool to summarize the
performance. We applied the Sum of Ranking Differences to summarize the comparison. Some aspects affecting SRD and its validation are: ties in input matrix, random ranking, and what gold standard to use. We discuss each one in the context of our application below.

(1) About ties. If the values in a column of the input matrix contains ties, the calculation of theoretical distribution of SRD needs special care (see \cite{Kollar-Hunek.Heberger2013}). For each algorithm, the test errors (or number of selected genes) for the ten data sets were ranked. For the same data set, two algorithms may have the same test error. However, they were not ranked together (see Table \ref{tab8}). Due to sample sizes being very different (see Table \ref{tab1}) for different cancer data sets, one misclassified sample contributes different percentages in test error for different cancer data sets.
Accordingly, the test error (calculated as average percentage of misclassified samples from 100 runs) is generally different from the ten data sets unless the errors are zero for all 100 runs. Therefore, it is not necessary to concern about ties in test error. Similarly, drastic difference in the total number of genes for different cancer data sets also leads to the number of selected genes showing no ties.

(2) About random ranking. We used the CRRN\_DNA software downloaded from the link given in \cite{Heberger2011} to calculate SRD and its theoretical distribution. According to \cite{Heberger2011}, discrete theoretical distribution is used in the case with 10 numbers to rank. That is, the recursive algorithm of \cite{Heberger2011} computes the probability of getting a permutation of $\{1, 2, \ldots, 10\}$ having SRD value less than a given number out of 10!= 3628800 permutations. The resulting probability is the exact probability (with no approximation). An alternative to the permutation test CRRN is cross-validation. As recommended in \cite{Kollar-Hunek.Heberger2013}, with number of data sets less than 14, leave-one-out cross-validation could be used to assess the uncertainties to the SRD values followed by Wilcoxon's matched pair test or sign test for significance.

(3) About gold standard.
The gold standard we used is the minimum of test errors from all algorithms in a comparison. This is because it is desired to have smaller test errors and less number of informative genes used in a classification algorithm. The algorithms with the smaller SRD values are closer to the ideal performance. Alternatively, one could use the maximum error and maximum number of genes as the gold standard. In that case, the algorithm whose  SRD value differs most with the gold standard will be the best one. In either case (using minimum or maximum as the gold standard), we believe the conclusion will be consistent. A third choice of the gold standard could be the average performance from ten data sets. Then the SRD is a nonparametric measure of how far each algorithm is away from the average performance. With this, however, we may not be able to tell whether an algorithm has the least test error because they are compared to the average.

}

Application to 10 cancer datasets reveals that none of the thresholding methods gives the best performance in all datasets. However, overall analyses {\modified with SRD} provide sufficient evidence to conclude that the order thresholding and hard thresholding lead to both improved classification accuracy and reduced number of selected variables. { \modified We can make this conclusion with confidence because the SRD values of the order thesholded and hard thresholded algorithms are significantly different from random ranking.}
The deep search helps improve performance significantly when it is used along with the hard thresholding and soft thresholding but does not seem to contribute positively in the case of order thresholding.

\section*{Acknowledgments}
This work was partially supported by a grant by Simons foundation (\#246077) to Haiyan Wang. {\modified We would also like to thank the anonymous reviewer whose comments have lead to a much improved version of this manuscript.}

\section*{Declaration of interest statement}
The authors do not have any potential competing interest.

\renewcommand{\baselinestretch}{1}

\bibliographystyle{chicago} 
\bibliography{references}

\begin{thebibliography}{}

\bibitem[\protect\citeauthoryear{Alizadeh, Eisen, Davis, and et~al}{Alizadeh
  et~al.}{2000}]{Alizadeh.et.al2000}
Alizadeh, A., M.~Eisen, R.~Davis, and et~al (2000).
\newblock Distinct types of diffuse large b-cell lymphoma identified by gene
  expression profiling.
\newblock {\em Nature\/}~{\em 403\/}(6769), 503--511.

\bibitem[\protect\citeauthoryear{Armstrong, Staunton, Silverman, and
  et~al}{Armstrong et~al.}{2002}]{Armstrong.et.al2002}
Armstrong, S.~A., J.~E. Staunton, L.~B. Silverman, and et~al (2002, January).
\newblock {MLL translocations specify a distinct gene expression profile that
  distinguishes a unique leukemia}.
\newblock {\em Nature Genetics\/}~{\em 30}, 41--47.

\bibitem[\protect\citeauthoryear{Beer, Kardia, Huang, and et~al}{Beer
  et~al.}{2002}]{Beer.et.al2002}
Beer, D.~G., S.~L. Kardia, C.-C. Huang, and et~al (2002).
\newblock {Gene-expression profiles predict survival of patients with lung
  adenocarcinoma}.
\newblock {\em Nature Medicine\/}~{\em 8\/}(8), 816--824.

\bibitem[\protect\citeauthoryear{Bhattacharjee, Richards, Staunton, and
  et~al}{Bhattacharjee et~al.}{2001}]{Bhattacharjee.et.al2001}
Bhattacharjee, A., W.~G. Richards, J.~Staunton, and et~al (2001).
\newblock Classification of human lung carcinomas by mrna expression profiling
  reveals distinct adenocarcinoma subclasses.
\newblock {\em Proceedings of the National Academy of Sciences\/}~{\em
  98\/}(24), 13790--13795.

\bibitem[\protect\citeauthoryear{{Bickel}}{{Bickel}}{1983}]{Bickel1983}
{Bickel}, P. (1983).
\newblock {Minimax estimation of the mean of a normal distribution subject to
  doing well at a point.}
\newblock {In \emph{Recent Advances in Statistics}, Pap. in Honor of H.
  Chernoff, Academic Press, New York, 511-528.}

\bibitem[\protect\citeauthoryear{Breiman}{Breiman}{1996}]{Breiman1996}
Breiman, L. (1996).
\newblock Bagging predictors.
\newblock {\em Machine Learning\/}~{\em 24\/}(2), 123--140.

\bibitem[\protect\citeauthoryear{Breiman}{Breiman}{2001}]{Breiman2001}
Breiman, L. (2001).
\newblock Random forests.
\newblock {\em Machine Learning\/}~{\em 45\/}(1), 5--32.

\bibitem[\protect\citeauthoryear{Chang and Lin}{Chang and
  Lin}{2011}]{Chang.Lin2011}
Chang, C.-C. and C.-J. Lin (2011).
\newblock Libsvm: A library for support vector machines.
\newblock {\em ACM Transactions on Intelligent Systems and Technology\/}~{\em
  2\/}(3), 1--27.

\bibitem[\protect\citeauthoryear{Chen and Guestrin}{Chen and
  Guestrin}{2016}]{Chen.Guestrin2016}
Chen, T. and C.~Guestrin (2016).
\newblock {Xgboost: A Scalable tree boosting system}.
\newblock {\em Proceeding of the 22nd ACM SIGKDD International Conference on
  Knowledge Discovery and Data Mining KDD'16\/}, 785--794.

\bibitem[\protect\citeauthoryear{Donoho and Johnstone}{Donoho and
  Johnstone}{1994}]{Donoho.Johnstone1994}
Donoho, D.~L. and J.~M. Johnstone (1994).
\newblock Ideal spatial adaptation by wavelet shrinkage.
\newblock {\em Biometrika\/}~{\em 81\/}(3), 425--455.

\bibitem[\protect\citeauthoryear{Fan}{Fan}{1996}]{Fan1996}
Fan, J. (1996).
\newblock Test of significance based on wavelet thresholding and
  {N}eyman{\textquoteright}s truncation.
\newblock {\em Journal of the American Statistical Association\/}~{\em
  91\/}(434), 674{\textendash}688.

\bibitem[\protect\citeauthoryear{Fan and Fan}{Fan and Fan}{2008}]{Fan.Fan2008}
Fan, J. and Y.~Fan (2008).
\newblock High-dimensional classification using features annealed independence
  rules.
\newblock {\em Annals of Statistics\/}~{\em 36}, 2605--2637.

\bibitem[\protect\citeauthoryear{Fan, Li, Sanz-Sole, Soria, Varona, and
  Verdera}{Fan et~al.}{2006}]{Fan.Li2006}
Fan, J., R.~Li, M.~Sanz-Sole, J.~Soria, J.~L. Varona, and J.~Verdera (2006).
\newblock Statistical challenges with high-dimensionality: Feature selection in
  knowledge discovery.
\newblock In {\em International congress of mathematicians}, Volume {III}, pp.\
   595--622. Providence, {RI}: American Mathematical Society.
\newblock Book, Section.

\bibitem[\protect\citeauthoryear{Freund and Schapire}{Freund and
  Schapire}{1997}]{Freund1997}
Freund, Y. and R.~E. Schapire (1997).
\newblock A decision-theoretic generalization of on-line learning and an
  application to boosting.
\newblock {\em Journal of Computer and System Sciences\/}~{\em 55\/}(1), 119 --
  139.

\bibitem[\protect\citeauthoryear{Friedman, Hastie, and Tibshirani}{Friedman
  et~al.}{2010}]{Friedman10regularizationpaths}
Friedman, J., T.~Hastie, and R.~Tibshirani (2010).
\newblock Regularization paths for generalized linear models via coordinate
  descent.
\newblock {\em Journal of Statistical Software\/}~{\em 33}, 1--22.

\bibitem[\protect\citeauthoryear{Golub, Slonim, Tamayo, and et~al}{Golub
  et~al.}{1999}]{Golub.et.al1999}
Golub, T., D.~Slonim, P.~Tamayo, and et~al (1999).
\newblock Molecular classification of cancer: class discovery and class
  prediction by gene expression monitoring.
\newblock {\em Science\/}~{\em 286}, 531--537.

\bibitem[\protect\citeauthoryear{H{\'e}berger}{H{\'e}berger}{2010}]{Heberger2010}
H{\'e}berger, K. (2010).
\newblock Sum of ranking differences compares methods or models fairly.
\newblock {\em Trends in Analytical Chemistry\/}~{\em 29\/}(1), 101--109.

\bibitem[\protect\citeauthoryear{H{\'e}berger and
  Koll{\'a}r-Hunek}{H{\'e}berger and Koll{\'a}r-Hunek}{2011}]{Heberger2011}
H{\'e}berger, K. and L.~Koll{\'a}r-Hunek (2011).
\newblock Sum of ranking differences for method discrimination and its
  validation: comparison of ranks with random numbers.
\newblock {\em Journal of Chemometrics\/}~{\em 25}, 151--158.

\bibitem[\protect\citeauthoryear{Johnstone and Silverman}{Johnstone and
  Silverman}{2004}]{Johnstone.Silverman2004}
Johnstone, I.~M. and B.~W. Silverman (2004).
\newblock Needles and straw in haystacks: empirical {B}ayes estimates of
  possibly sparse sequences.
\newblock {\em Annals of Statistics\/}~{\em 32}, 1594--1649.

\bibitem[\protect\citeauthoryear{Khan, Wei, Ringner, and et~al}{Khan
  et~al.}{2001}]{Khan.et.al2001}
Khan, J., J.~S. Wei, M.~Ringner, and et~al (2001).
\newblock Classification and diagnostic prediction of cancers using gene
  expression profiling and artificial neural networks.
\newblock {\em Nature Medicine\/}~{\em 7\/}(6), 673--679.

\bibitem[\protect\citeauthoryear{Kim and Akritas}{Kim and
  Akritas}{2010}]{Kim.Akritas2010}
Kim, M. and M.~Akritas (2010).
\newblock Order thresholding.
\newblock {\em Annals of Statistics\/}~{\em 38\/}(4), 2314--2350.

\bibitem[\protect\citeauthoryear{Kollár-Hunek and Héberger}{Kollár-Hunek and
  Héberger}{2013}]{Kollar-Hunek.Heberger2013}
Kollár-Hunek, K. and K.~Héberger (2013).
\newblock Method and model comparison by sum of ranking differences in cases of
  repeated observations (ties).
\newblock {\em Chemometrics and Intelligent Laboratory Systems\/}~{\em 127},
  139--146.

\bibitem[\protect\citeauthoryear{Ng and Jordan}{Ng and
  Jordan}{2002}]{Ng.Jordan2002}
Ng, A. and M.~Jordan (2002).
\newblock {On discriminative vs. generative classifiers: A comparsion of
  logistic regression and \protect{Naive Bayes} }.
\newblock {\em Advances in Neural Iformation Processing Systems\/}~{\em 14},
  841--848.

\bibitem[\protect\citeauthoryear{Perou, Sorlie, Eisen, and et~al}{Perou
  et~al.}{2000}]{Perou.et.al2000}
Perou, C.~M., T.~Sorlie, M.~B. Eisen, and et~al (2000).
\newblock Molecular portraits of human breast tumours.
\newblock {\em Nature\/}~{\em 406}, 747--752.

\bibitem[\protect\citeauthoryear{Ramaswamy, Tamayo, Rifkin, and
  et~al}{Ramaswamy et~al.}{2001}]{Ramaswamy.et.al2001}
Ramaswamy, S., P.~Tamayo, R.~Rifkin, and et~al (2001).
\newblock Multiclass cancer diagnosis using tumor gene expression signatures.
\newblock {\em Proceedings of the National Academy of Sciences\/}~{\em
  98\/}(26), 15149--15154.

\bibitem[\protect\citeauthoryear{Su, Welsh, Sapinoso, and et~al}{Su
  et~al.}{2001}]{Su.et.al2001}
Su, A., J.~Welsh, L.~Sapinoso, and et~al (2001).
\newblock Molecular classification of human carcinomas by use of gene
  expression signatures.
\newblock {\em Cancer Research\/}~{\em 61\/}(20), 7388--7393.

\bibitem[\protect\citeauthoryear{Tan, Naiman, Xu, Winslow, and Geman}{Tan
  et~al.}{2005}]{Tan.et.al2005}
Tan, A.~C., D.~Q. Naiman, L.~Xu, R.~L. Winslow, and D.~Geman (2005).
\newblock Simple decision rules for classifying human cancers from gene
  expression profiles.
\newblock {\em Bioinformatics\/}~{\em 21\/}(20), 3896--3904.

\bibitem[\protect\citeauthoryear{Tibshirani, Hastie, Narasimhan, and
  Chu}{Tibshirani et~al.}{2002}]{Tibshirani.et.al2002}
Tibshirani, R., T.~Hastie, B.~Narasimhan, and G.~Chu (2002).
\newblock Diagnosis of multiple cancer types by shrunken centroids of gene
  expression.
\newblock {\em Proceedings of the National Academy of Sciences\/}~{\em
  99\/}(10), 6567--6572.

\bibitem[\protect\citeauthoryear{Yeoh, Ross, Shurtleff, and et~al}{Yeoh
  et~al.}{2002}]{Yeoh.et.al2002}
Yeoh, E., M.~E. Ross, S.~A. Shurtleff, and et~al (2002).
\newblock Classification, subtype discovery, and prediction of outcome in
  pediatric acute lymphoblastic leukemia by gene expression profiling.
\newblock {\em Cancer Cell\/}~{\em 1\/}(2), 133--143.

\bibitem[\protect\citeauthoryear{Zhang, Wang, Dai, Chen, and Yuan}{Zhang
  et~al.}{2012}]{Zhang.Wang2012}
Zhang, H., H.~Wang, Z.~Dai, M.-s. Chen, and Z.~Yuan (2012).
\newblock Improving accuracy for cancer classification with a new algorithm for
  genes selection.
\newblock {\em BMC Bioinformatics\/}~{\em 13}, 1--20.

\bibitem[\protect\citeauthoryear{Zhang, Ahn, and Lin}{Zhang
  et~al.}{2006}]{Zhang.et.al2006}
Zhang, H.~H., J.~Ahn, and X.~Lin (2006).
\newblock Gene selection using support vector machines with nonconvex penalty.
\newblock {\em Bioinformatics\/}~{\em 22}, 88--95.

\end{thebibliography}


%

\section*{Tables}
\begin{table}[h!]
 \begin{center}
 \caption{Illustration of potential problem of thresholding parameter
 estimate in PAM. This is obtained for Leukemia2 data using pamr.cv from R package pamr
 with the seed of random number generation set to  set.seed=100 in R
 2.15.0. The number of genes survived soft thresholding corresponding
 to the smallest cv error could be drastically different from that
 corresponding to the second smallest cv error.}
 \vspace{.1in}
 \begin{tabular}{|l|lll|lll|}
   \hline
 &\multicolumn{3}{c|}{Parameter with smallest CV error}& \multicolumn{3}{c|}{Parameter with 2nd smallest CV
 error}\\
   \cline{2-7}
 & threshold & n.genes &  CV error & threshold & n.genes &
  CV error \\
   \hline
  run 1 & 0.418878 & 10283 & 5 & 7.539809 & 26 & 6 \\
  run 2 & 1.256635 & 6127 & 4 & 7.539809 & 26 & 6 \\
  run 3 & 0.418878 & 10283 & 4 & 7.12093 & 30 & 5 \\
  run 4 & 0.837757 & 7959 & 3 & 6.283174 & 77 & 4 \\
  run 5 & 6.283174 & 77 & 5 & 6.702052 & 52 & 6 \\
  run 6 & 1.675513 & 4735 & 5 & 6.702052 & 52 & 6 \\
  run 7 & 1.256635 & 6127 & 4 & 7.539809 & 26 & 5 \\
  run 8 & 7.539809 & 26 & 6 & 7.958687 & 18 & 7 \\
  run 9 & 6.702052 & 52 & 4 & 7.958687 & 18 & 5 \\
  run 10 & 1.256635 & 6127 & 5 & 0.418878 & 10283 & 6 \\
    \hline
 \end{tabular}
  \label{Table}
 \end{center}
 \end{table}


\begin{table}[H]
\small
\begin{center}
\caption{Summary of data sets used.}
\begin{tabular}{|l|lccccl|}
  \hline
Dataset & Platform & No. of & No. of & \multicolumn{2}{c}{No. of samples}& Reference \\
abbreviation & & classes & genes &Training &Testing &  \\ \hline
SRBCT& cDNA &4 &  2308 & 63 &  20 & \cite{Khan.et.al2001} \\
Breast& Affy &5 &  9216 & 54 &  30 & \cite{Perou.et.al2000} \\
Cancers& Affy &11 &  12533 & 100 &  74 & \cite{Su.et.al2001} \\
DLBCL& cDNA &6 &  4026 & 58 &  30 & \cite{Alizadeh.et.al2000} \\
GCM& Affy &14 &  16063 & 144 &  46 & \cite{Ramaswamy.et.al2001} \\
Leukemia1& Affy &3 &  7129 & 38 &  34 & \cite{Golub.et.al1999} \\
Leukemia2& Affy &3 &  12582 & 57 &  15 & \cite{Armstrong.et.al2002} \\
Leukemia3& Affy &7 &  12558 & 215 &  112 & \cite{Yeoh.et.al2002} \\
Lung1& Affy &3 &  7129 & 64 &  32 & \cite{Beer.et.al2002} \\
Lung2& Affy &5 &  12600 & 136 &  67 & \cite{Bhattacharjee.et.al2001} \\
\hline
\end{tabular}
 \label{tab1}
\end{center}
\end{table}

\begin{table}[H]
\small
\caption{{\small The SRD of mean test errors for the three thresholding methods. The column 'Min' refers to the smallest error from the three methods, 'rank' is the rank of error within the same {\modified column}, and 'diff' refers to the absolute difference in rank compared to the golden standard.} }
\centering
\begin{tabular}{|r|c|cc|ccc|ccc|ccc|}
  \hline
& \multicolumn{1}{c|}{} &\multicolumn{2}{c|}{Golden standard}& \multicolumn{3}{c|}{STh}&\multicolumn{3}{c|}{OTh}&\multicolumn{3}{c|}{HTh}\\
 &  &Min& rank& error& rank& diff& error& rank& diff& error& rank& diff\\
  \hline
 &Lung2&  0.00&   1&  1.33&   2&  1&  0   &1  &0  &2.7&   2&  1\\
&DLBCL&  0.97&   2&  8.7&    6&  4&  1.63&   2&  0&  0.97&   1&  1\\
&Leukemia3&  1.11&   3&  1.11&   1&  2&  5.01&   3&  0&  4.26&   3&  0\\
&Leukemia1&  3.00&   4&  3&  3&  1&  11.5&   6&  2&  11.79&  7&  3\\
Data&SRBCT&  5.00&   5&  5&  4&  1&  5.2&    4&  1&  5&  4&  1\\
sets&Breast& 5.70&   6&  6.23&   5&  1&  5.7&    5&  1&  7.93&   5&  1\\
&Leukemia2&  11.53&  7&  13.73&  8&  1&  13.2&   7&  0&  11.53&  6&  1\\
&Cancers&    12.05&  8&  12.05&  7&  1&  16.42&  8&  0&  16.35&  8&  0\\
&Lung1&  18.62&  9&  21.84&  9&  0&  18.75&  9&  0&  18.62&  9&  0\\
&GCM&    44.00&  10& 44& 10& 0&  51.7&   10& 0&  52.46&  10& 0\\
   \hline
sum& &&&&&               12&&&           4&&&            8\\
   \hline
\end{tabular}
\label{tab8}
\end{table}

\begin{table}[H]
\small
\begin{center}
\caption{Average number of informative genes based on 100 runs for each thresholding method. The value in parenthesis is the standard error.}
\vspace{.1in}
\begin{tabular}{|l|ccc|}
  \hline
 & STh & HTh & OTh \\
  \hline
  Lung1 & 50(6.0) & 134(42.7) & 87(16.5) \\
  SRBCT & 94(8.0) & 36(1.0) & 32(.9) \\
  Leukemia1 & 111(50.0) & 149(18.4) & 139(12.0) \\
  Cancers & 1111(37.7) & 1548(66.4) & 1469(39.6) \\
  Lung2 & 1911(169.3) & 3610(88.1) & 2106(38.0) \\
  GCM & 2010(89.9) & 3716(212.9) & 2931(33.9) \\
  Breast & 3317(152.3) & 1494(132.5) & 679(43.9) \\
  DLBCL & 3483(63.9) & 716(55.4) & 360(8.7) \\
  Leukemia2 & 5389(414.9) & 1492(258.1) & 327(47.5) \\
  Leukemia3 & 8637(208.9) & 2073(254.5) & 1156(38.6) \\
   \hline
   overall average&2611&1497&929\\
   \hline
\end{tabular}
\label{tab3}
\end{center}
\end{table}


\begin{table}[H]
\scriptsize
\caption{The percent of mean misclassification error for test samples and average number of informative genes based on 100 runs for each
thresholding method with and without the deep search algorithm. {\modified STh, HTh, OTh: without deep search; STh2, HTh2, OTh2: with deep search. }}
\begin{tabular}{|l|llll|llll|llll|}
  \hline
  &\multicolumn{2}{c}{STh}&\multicolumn{2}{c|}{STh2}&\multicolumn{2}{c}{HTh}&\multicolumn{2}{c|}{HTh2}&\multicolumn{2}{c}{OTh}&\multicolumn{2}{c|}{OTh2}\\
 &error&n.genes&error&n.genes&error&n.genes&error&n.genes&error&n.genes&error&n.genes\\
  \hline
  SRBCT &5&94&5&18&5&36&5&26&5.2&32&5&30 \\
  Breast &6.23&3317&6.33&2266&7.93&1494&5.57&549&5.7&679&6.87&371 \\
  Cancers &12.05&1111&12.31&956&16.35&1548&15.65&1631&16.42&1469&16.89&1360 \\
  DLBCL &8.7&3483&8.97&3399&0.97&716&0.83&491&1.63&360&1.83&250 \\
  GCM &44&2010&43.87&1692&52.46&3716&54.11&3709&51.7&2931&51.61&3009 \\
  Leukemia1 &3&111&3.09&41&11.79&149&11.32&190&11.5&139&9.68&169 \\
  Leukemia2 &13.73&5389&6.73&2236&11.53&1492&7.6&208&13.2&327&8.13&109 \\
  Leukemia3 &1.11&8637&3.01&4606&4.26&2073&4.85&1943&5.01&1156&5.07&1020 \\
  Lung1 &21.84 &50&21.62&13&18.62 &134&18.47&48&18.75 &87&18.69&91 \\
  Lung2 &1.33&1911&0.69&717&2.7&3610&2.06&3290&0&2106&0.01&2083 \\
   \hline
 \end{tabular}
\label{tab4}
\end{table}

\newpage
\listoffigures
\section*{Figures}
\begin{figure}[H]
\caption{{Test error of OTh and HTh versus STh for SRBCT and Breast cancer data. \small The plotting symbol H (in red) is
for HTh and O (in black) is for OTh. The numbers used in the plot
are the frequencies of test errors out of 100 runs and the table gives
a summary of the test errors in percentage.}}
 \label{I1}
 \centering
\footnotesize
\begin{tabular}{cc}
\includegraphics[
width=.47\linewidth, angle=0]{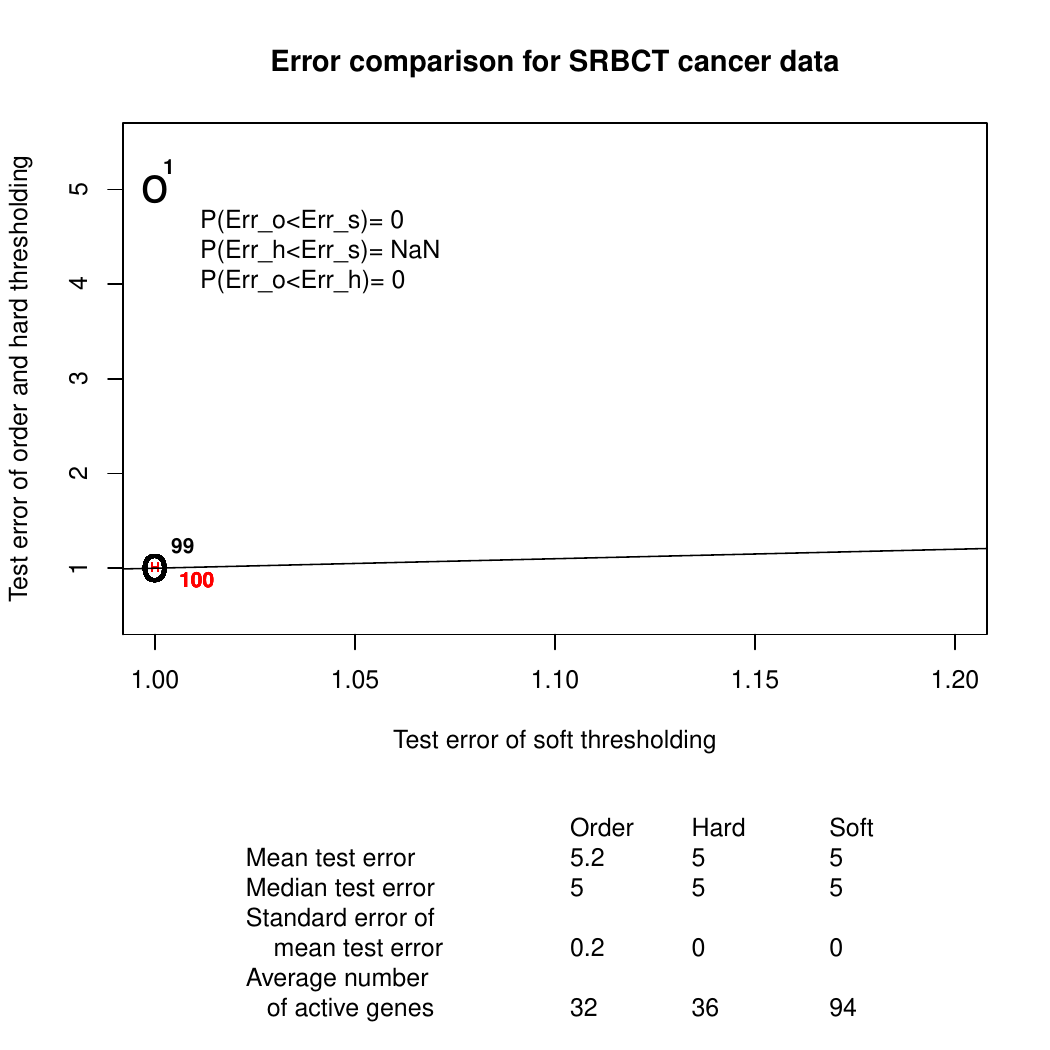}
&
\includegraphics[
width=0.47\linewidth, angle=0]{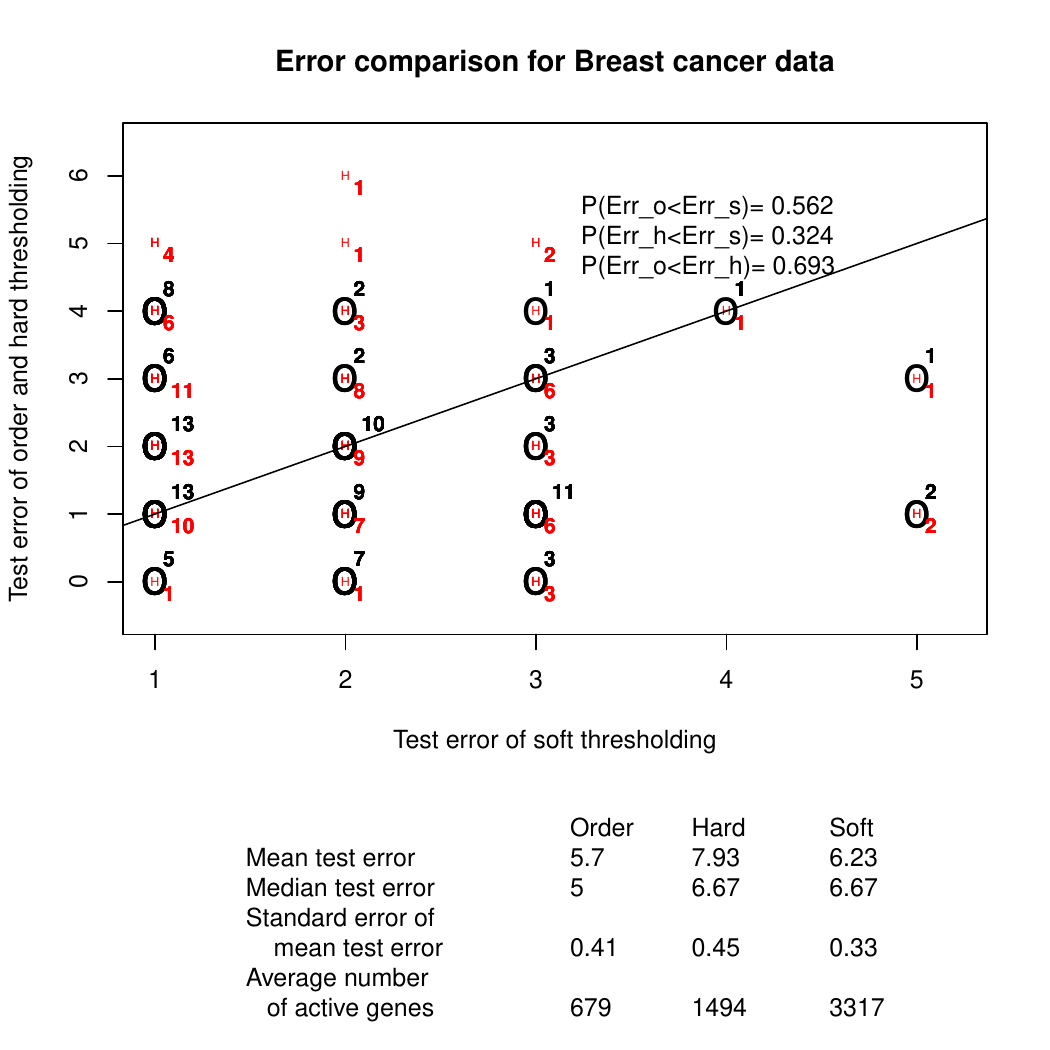}
 \\
(a) SRBCT dataset analysis & (b)Breast cancer dataset
analysis \\
\end{tabular}
\end{figure}

\newpage
\appendix

\renewcommand{\theequation}{S.\arabic{equation}}
 \setcounter{equation}{0}  

\renewcommand\thefigure{S.\arabic{figure}}
\setcounter{figure}{0}

 \setcounter{section}{0}\renewcommand\thesection {A}

\bigskip\noindent\textsf{\Large Supplementary Material for article: ``Different thresholding methods on Nearest Shrunken Centroid algorithm" by Mohammad Omar Sahtout, Haiyan Wang, and Santosh Ghimire.}

\begin{figure}[htbp]
\caption{Test error of OTh and HTh versus STh for Cancers and DLBCL data. {\modified \scriptsize \small The plotting symbol H (in red) is
for HTh and O (in black) is for OTh. The numbers used in the plot
are the frequencies of test errors out of 100 runs and the table gives
a summary of the test errors in percentage.}}
 \label{I2}
 \centering
\footnotesize
\begin{tabular}{cc}
\includegraphics[height=.47\linewidth,width=.47\linewidth, angle=0]{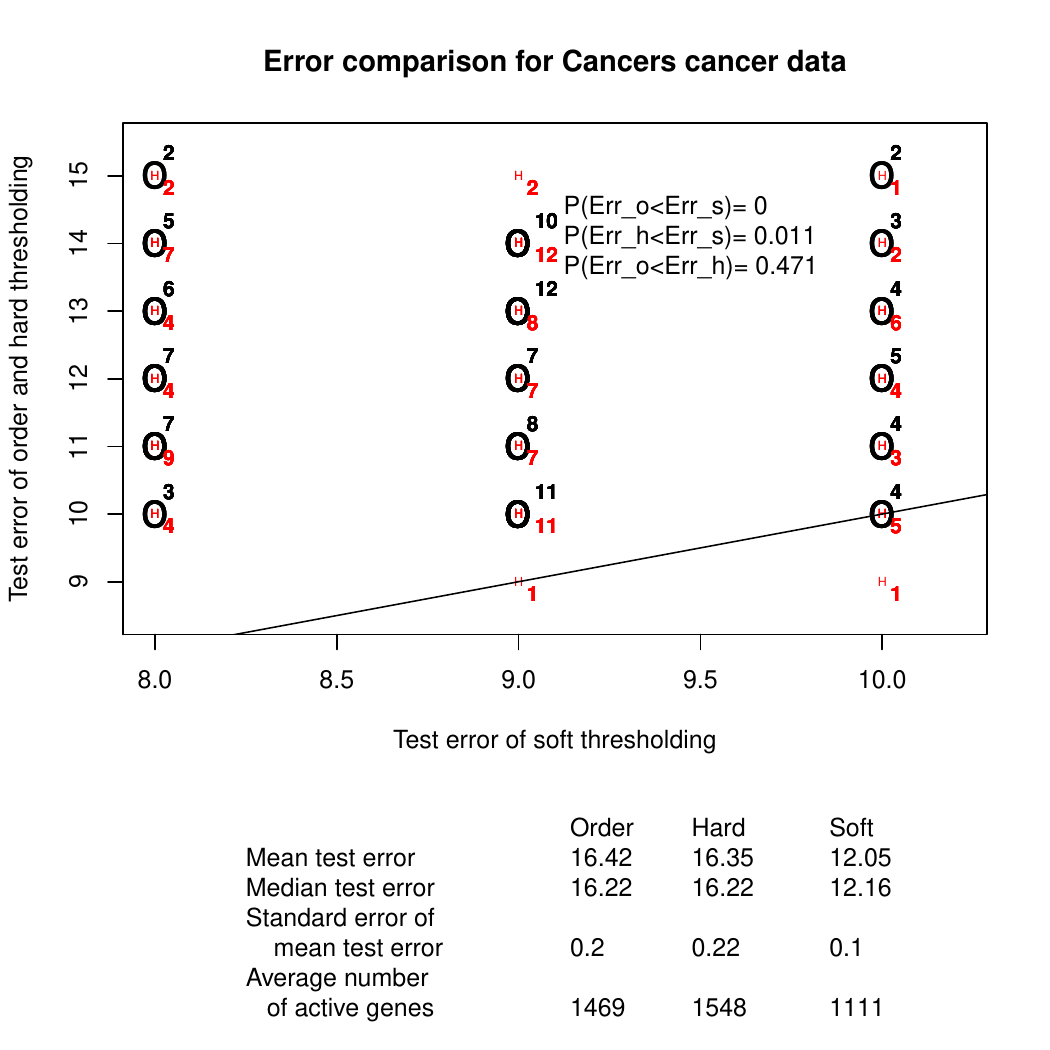}
&
\includegraphics[height=.47\linewidth,width=.47\linewidth, angle=0]{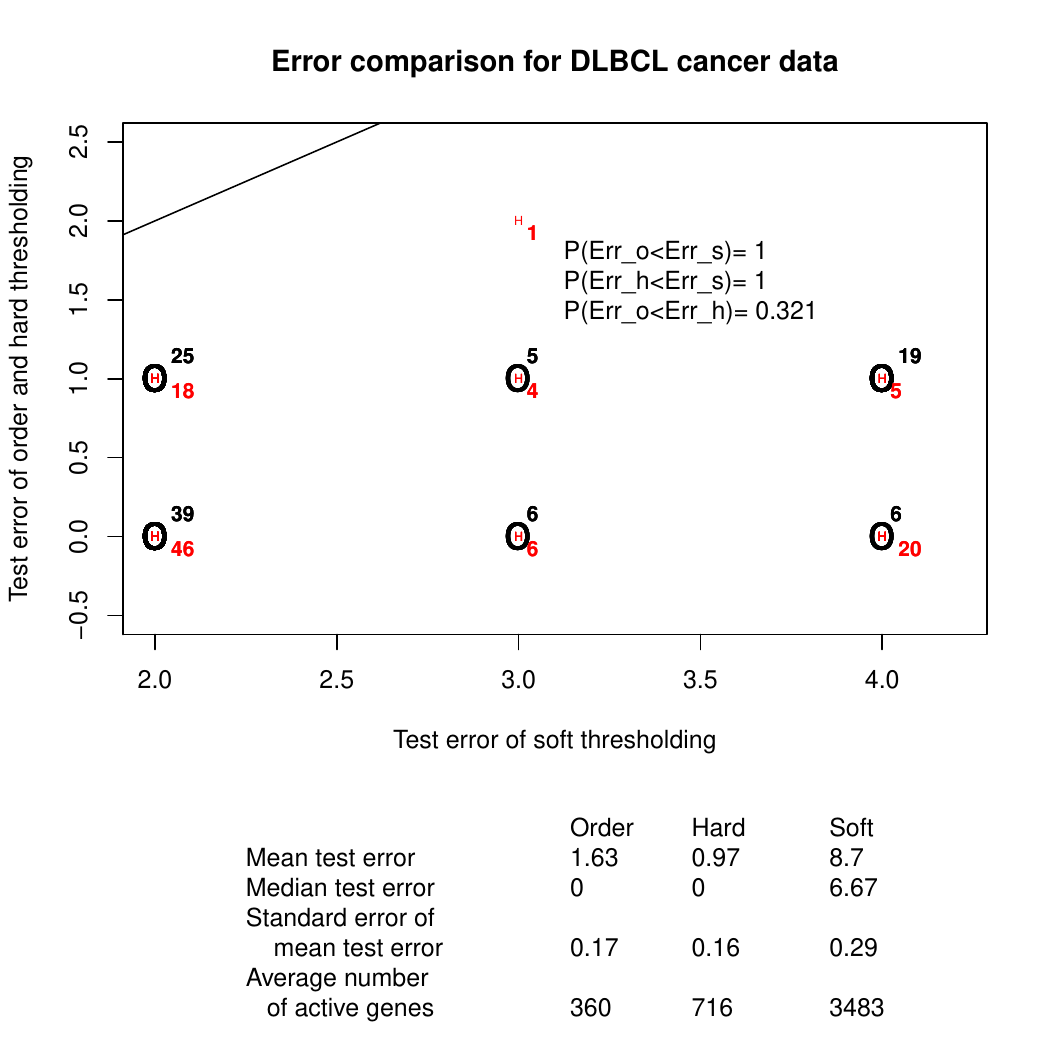}
 \\
(a) Cancer dataset analysis & (b) DLBC dataset analysis \\
\end{tabular}
\end{figure}

\begin{figure}[htbp]
\caption{Test error of OTh and HTh versus STh for GCM data. {\modified \scriptsize \small The plotting symbol H (in red) is
for HTh and O (in black) is for OTh. The numbers used in the plot
are the frequencies of test errors out of 100 runs and the table gives
a summary of the test errors in percentage.}}
 \label{I3}
 \centering
\includegraphics[
width=0.75\linewidth, angle=0]{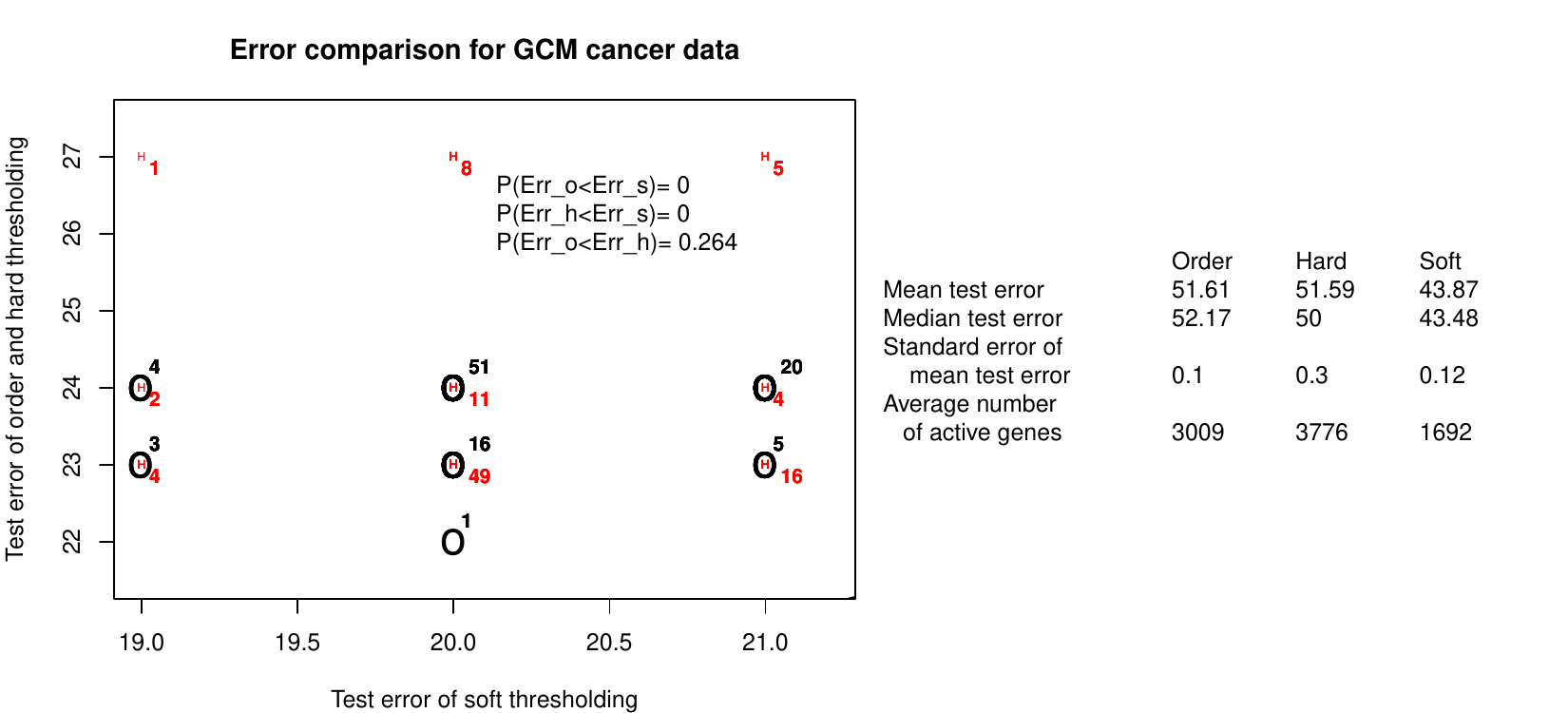}\\
GCM dataset analysis.
\end{figure}

\begin{figure}[htbp]
\caption{Test error of OTh and HTh versus STh for Leukemia1 and Leukemia2 data. {\modified \scriptsize \small The plotting symbol H (in red) is
for HTh and O (in black) is for OTh. The numbers used in the plot
are the frequencies of test errors out of 100 runs and the table gives
a summary of the test errors in percentage.}}
 \label{I4}
 \centering
\footnotesize
\begin{tabular}{cc}
\includegraphics[height=.5\linewidth,width=.5\linewidth, angle=0]{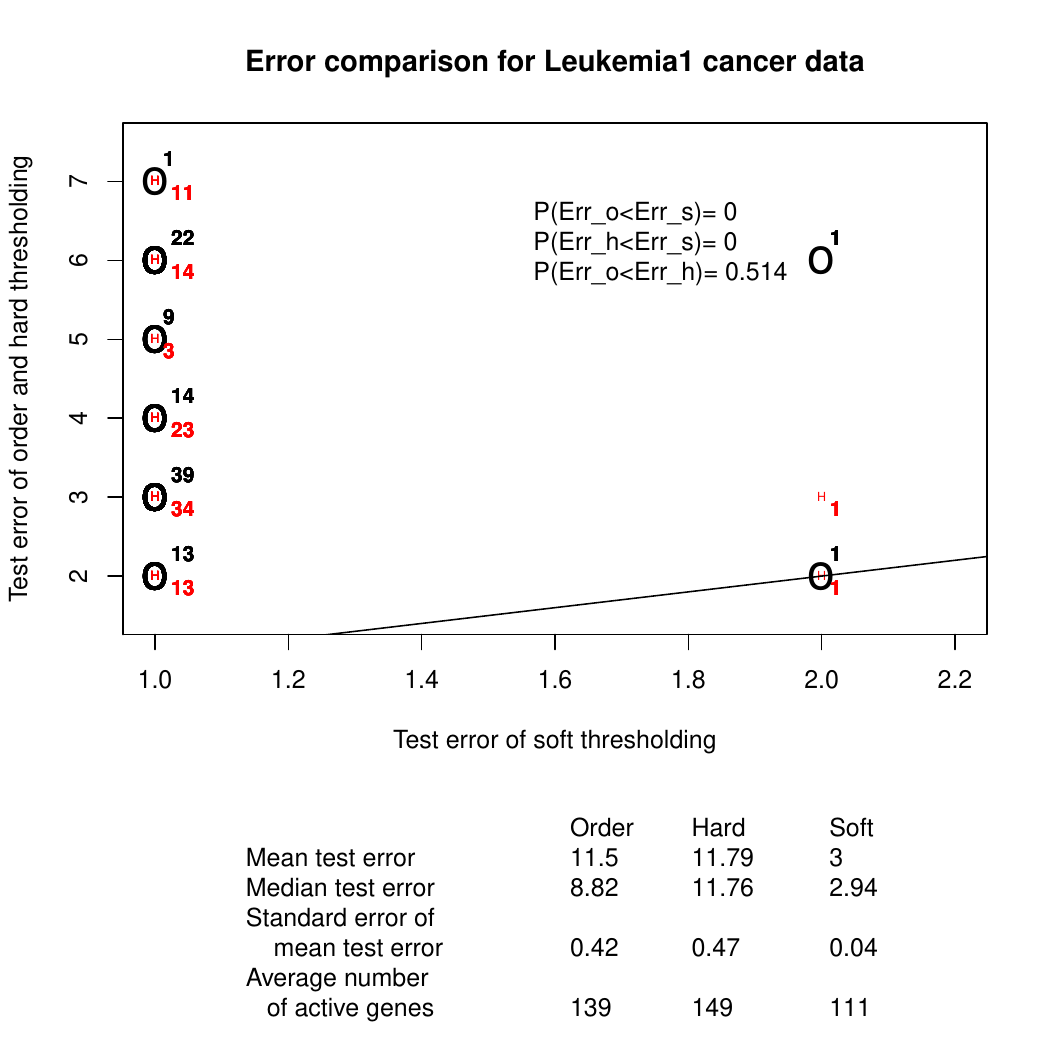}
&
\includegraphics[height=.5\linewidth,width=.5\linewidth, angle=0]{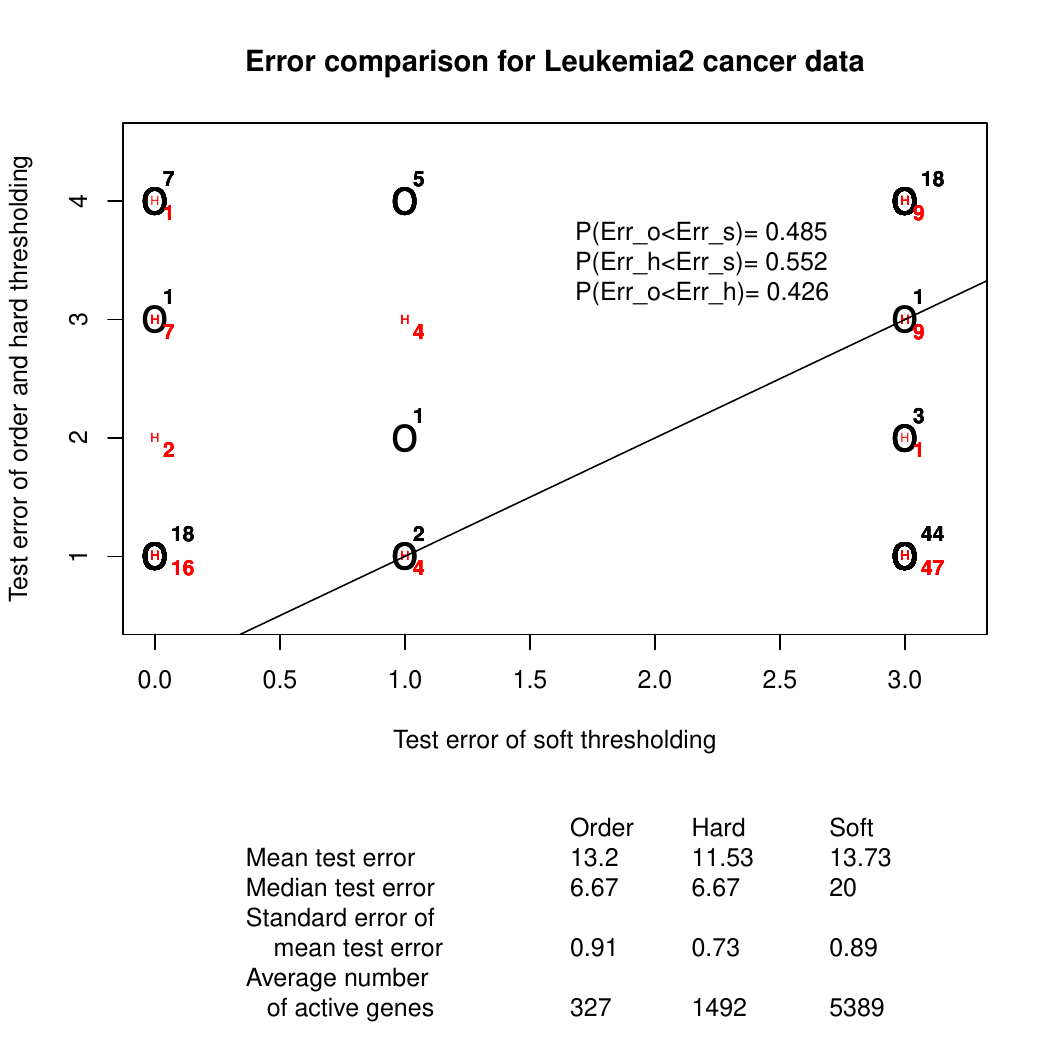}
 \\
(a) Leukemia1 cancer dataset  analysis & (b) Leukemia2 cancer dataset
 analysis \\
\end{tabular}
\end{figure}

\begin{figure}[htbp]
\caption{Test error of OTh and HTh versus STh for Leukemia3 data. {\modified \scriptsize \small The plotting symbol H (in red) is
for HTh and O (in black) is for OTh. The numbers used in the plot
are the frequencies of test errors out of 100 runs and the table gives
a summary of the test errors in percentage.}}
 \label{I5}
 \centering
\includegraphics[width=0.75\linewidth, angle=0]{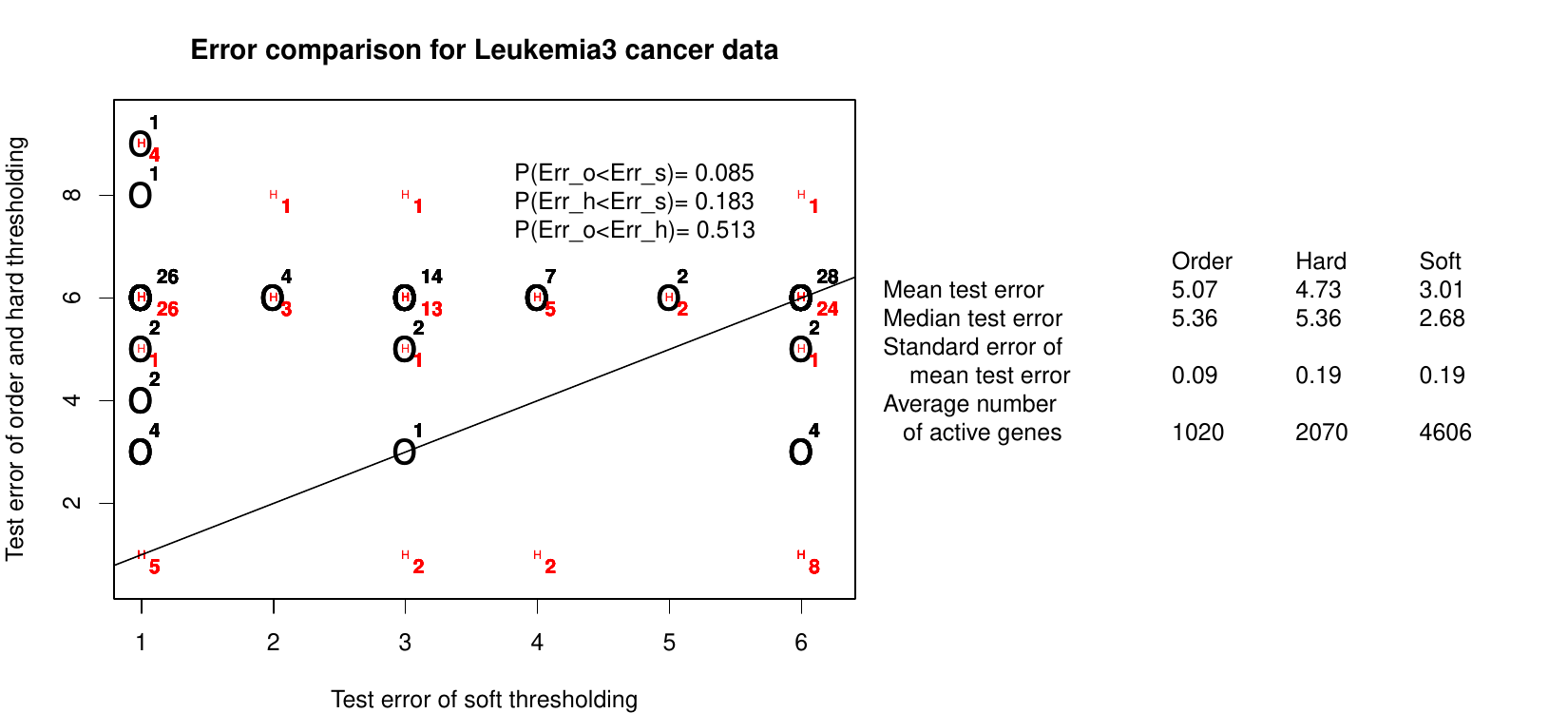}
\end{figure}

\begin{figure}[htbp]
\caption{Test error of OTh and HTh versus STh for Lung1 and Lung2 data. {\modified \scriptsize \small The plotting symbol H (in red) is
for HTh and O (in black) is for OTh. The numbers used in the plot
are the frequencies of test errors out of 100 runs and the table gives
a summary of the test errors in percentage.}}
 \label{I6}
 \centering
\footnotesize
\begin{tabular}{cc}
\includegraphics[height=.5\linewidth,width=.5\linewidth, angle=0]{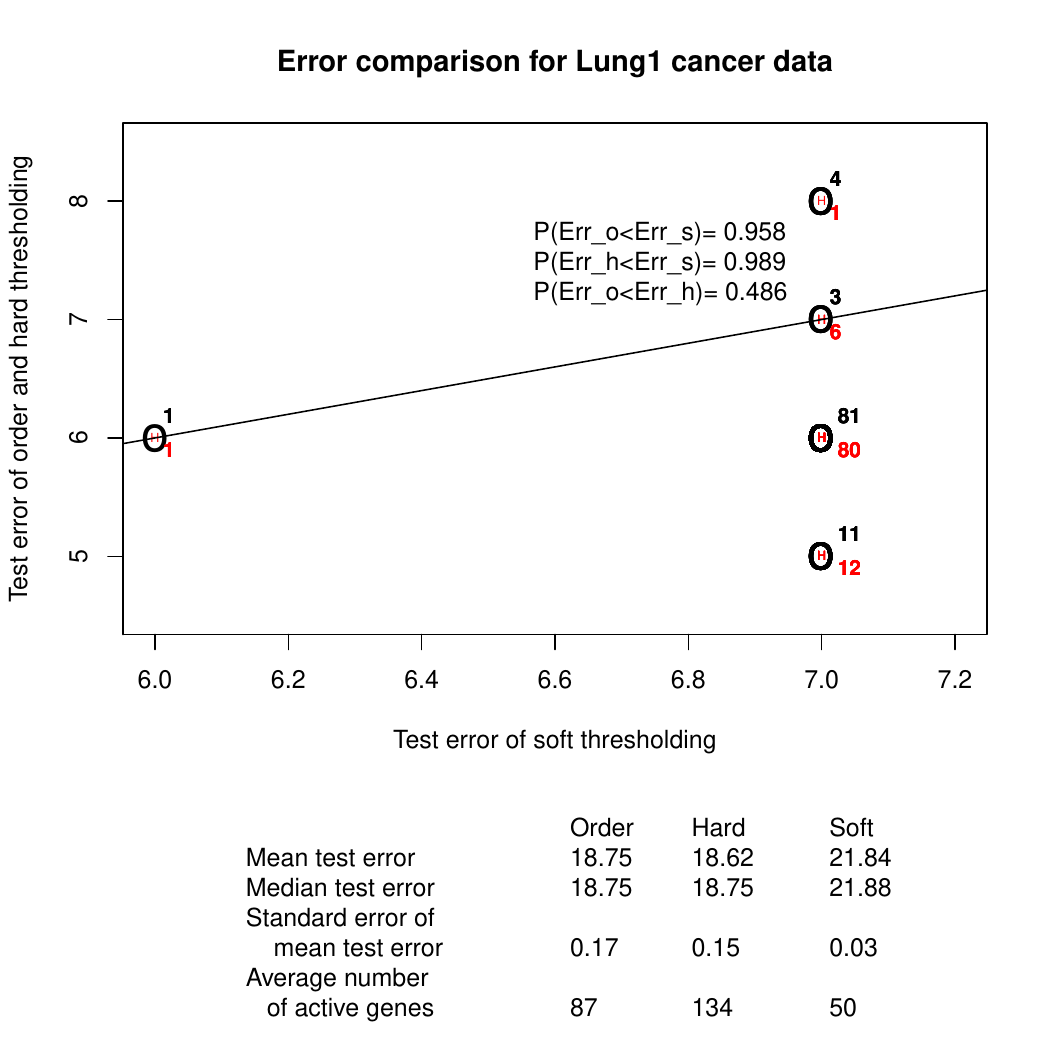}
&
\includegraphics[height=.5\linewidth,width=.5\linewidth, angle=0]{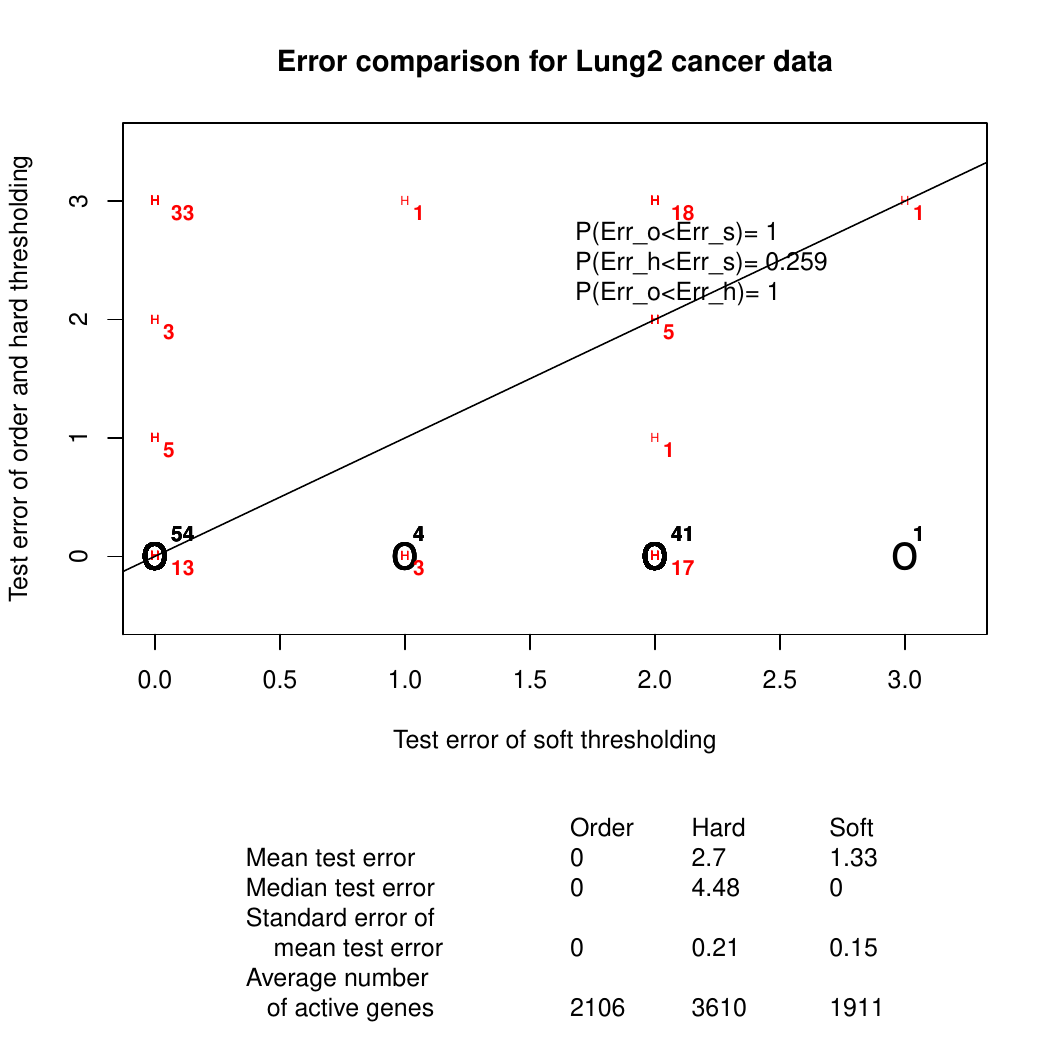}
 \\
(a) Lung1 cancer dataset analysis & (b) Lung2 cancer dataset
 analysis \\
\end{tabular}
\end{figure}

\medskip 
{\modified
This supplementary material presents Figures \ref{I2} - \ref{I6} mentioned in the article.
}

\end{document}